\documentclass[twocolumn,10pt]{jmlr}

\jmlrproceedings{}{}

\usepackage{lmodern}
\usepackage{booktabs}
\usepackage{multirow}
\usepackage{siunitx}
\usepackage{colortbl}
\usepackage{array}
\usepackage{longtable}
\usepackage[switch]{lineno}
\usepackage{soul}
\usepackage{color} 

\definecolor{agrow}{rgb}{0.87,0.93,1.0}

\newcommand{\equal}[1]{{\hypersetup{linkcolor=black}\thanks{#1}}}

\title[\texttt{AG-JEPA} for Neurodegenerative Representation Learning]{AG-JEPA: Anatomically Guided Joint Embedding Predictive Architecture for Neurodegenerative Representation Learning}

\author{%
  \Name{Azad Singh}\equal{These authors contributed equally.} \Email{t-singhazad@microsoft.com
}\\
  \Name{Ayantika Das}\footnotemark[1]\Email{t-aydas@microsoft.com}\\
  \Name{Amog Rao} \Email{t-amograo@microsoft.com}\\
  \Name{Aiman Farooq} \Email{t-afarooq@microsoft.com}\\
  \Name{Anirban Porya} \Email{t-aniporya@microsoft.com}\\
  \Name{Ananya Shukla} \Email{t-anashukla@microsoft.com}\\
  \Name{Sowbhagya Suresh}\Email{v-sosuresh@microsoft.com}\\
  \Name{Srinivasan Iyengar} \Email{sriyengar@microsoft.com}\\
  \Name{Mohit Jain} \Email{mohja@microsoft.com}\\
  \Name{Mercy Prasanna Ranjit} \Email{meranjit@microsoft.com}\\
  \addr Microsoft Research India
}

\hypersetup{
  pdftitle={AG-JEPA: Anatomically Guided Joint Embedding Predictive Architecture for Neurodegenerative Representation Learning},
  pdfauthor={Azad Singh, Ayantika Das, Amog Rao, Aiman Farooq, Anirban Porya, Ananya Shukla, Sowbhagya Suresh, Srinivasan Iyengar, Mohit Jain, Mercy Prasanna Ranjit}
}

\begin{document}
\maketitle
\begin{abstract}

Brain MRI can reveal early signs of Alzheimer’s disease (AD) before measurable cognitive decline becomes apparent. However, during the transition from cognitively normal (CN) to mild cognitive impairment (MCI), the disease-related signals are often subtle and confined to only a few brain regions. Existing Joint Embedding Predictive Architecture (JEPA) models learn general structural representations but do not explicitly capture the anatomical regions affected in AD. We propose Anatomically Guided JEPA (AG-JEPA), which introduces anatomical tokens that interact with patch tokens during pretraining to learn region-specific representations from brain regions relevant to neurodegeneration. Building on this representation space, we further propose \texttt{Prog-AG-JEPA}, which models disease progression trajectory as a conditioned latent displacement within this space. 
We evaluate \texttt{AG-JEPA} across five downstream tasks spanning Alzheimer's disease diagnosis and progression prediction. \texttt{AG-JEPA} improves diagnosis and MCI prognosis, achieving an 8.4 percentage point AUROC gain over the strongest baseline on CN/MCI classification and a 7.1 percentage point gain on out-of-distribution CN/AD classification on the MR-RATE dataset, highlighting its effectiveness across both early-stage detection and cross-cohort generalization. For disease progression modeling, \texttt{Prog-AG-JEPA} consistently outperforms all baselines on future diagnosis and brain volume change prediction, demonstrating the value of anatomically grounded neurodegeneration representations for diagnosis and prognosis. \end{abstract}

\begin{keywords}
Alzheimer's Disease, Mild Cognitive Impairment, MRI, JEPA, Neuroimaging Foundation Models
\end{keywords}

\paragraph*{Data and Code Availability.}
This study uses public datasets under their respective data-use agreements. Details of the datasets used are provided in Section~\ref{sec:exp}. At present, we do not release the code. \paragraph*{Institutional Review Board (IRB).} 
The use of de-identified datasets was reviewed by our home institution and determined not to constitute human subjects research under institutional policy.

\section{Introduction}
\label{sec:intro}

Alzheimer's disease (AD) is a progressive brain disorder that gradually worsens over time. It progresses from cognitively normal (CN) status through Mild Cognitive Impairment (MCI) and eventually to mild, moderate, and severe stages of AD dementia. The MCI stage is particularly critical, as timely interventions at this stage may slow or stabilize cognitive decline and potentially delay progression to AD dementia~\citep{petersen2009adni,niaalz2023}. Additionally, disease progression during MCI varies substantially across individuals: some remain stable, while others progress to AD. This heterogeneity makes the early identification and stratification of individuals with MCI especially important.

As AD progresses, brain tissue is lost in specific regions, a process known as atrophy. Structural MRI is typically used to visualize these changes and stage patients along the AD trajectory. During early MCI, neurodegenerative changes are largely confined to a small number of structures, including atrophy of the hippocampus and entorhinal cortex and expansion of adjacent ventricular regions~\citep{wen2020cnnad,petersen2009adni}. As the disease advances, these changes become more widespread across the brain~\citep{wen2020cnnad,petersen2009adni}.Because these changes are subtle and overlap with normal ageing, distinguishing CN from MCI remains a challenging task~\citep{wen2020cnnad,niaalz2023}.

Neuroimaging foundation models have advanced through latent predictive pre-training. Models such as \texttt{Neuro-JEPA}~\citep{huang2026neurojepa} and \texttt{NeuroVFM}~\citep{kondepudi2026neurovfm}, trained with the Joint Embedding Predictive Architecture (JEPA) objective, demonstrate strong performance across a range of downstream tasks. However, the objective operates at the patch level by predicting latent representations of masked regions and thus lacks anatomical awareness in the representational space~\citep{assran2023jepa}. While this is effective for learning broad structural representations, it may be limiting for tasks such as CN versus MCI classification~\citep{wen2020cnnad,niaalz2023}, where the discriminative signal is subtle and confined to few brain regions.

To address this limitation, we propose \texttt{AG-JEPA}, which augments standard patch representations in JEPA with anatomical tokens corresponding to regions that exhibit the earliest structural changes in Alzheimer's disease. However, adding such tokens alone does not guarantee that they encode clinically meaningful information about each structure. We therefore impose three constraints. \texttt{First}, the anatomical tokens must predict the masked image content without access to the unmasked patches, encouraging each token to encode the current condition of its region, rather than a generic regional template. \texttt{Second}, the visible patches must predict the anatomical tokens; together, these two objectives require the patch and token streams to remain mutually predictable, keeping the tokens tied to the observed image. \texttt{Third}, each token is aligned with the pooled representation of its anatomical region, so that it captures information about that specific region rather than being interchangeable across regions. Together, these constraints encourage each token to encode the condition of its structure in a given scan, allowing subtle, focal changes to be attributed to the region that produced them.

Building on this representation, we introduce \texttt{Prog-AG-JEPA}, which models longitudinal disease progression as displacement in the latent space. As the \texttt{AG-JEPA}'s latent space is already organized by region, a displacement within it directly expresses how each region changes over time, recovering region-level progression without ever reconstructing an image.

\textbf{Key Contributions:}
\begin{enumerate}
\setlength{\itemsep}{1pt} \item \textbf{An anatomically grounded pretraining objective.} We introduce \texttt{AG-JEPA}, a JEPA formulation that augments patch representations with anatomical tokens and jointly constrains them through masked-content prediction, token reconstruction, and region-specific alignment, yielding representations that remain interpretable at the region level.

\item \textbf{Latent modeling of longitudinal progression.} We introduce \texttt{Prog-AG-JEPA}, a progression head that models disease progression as displacements in the frozen \texttt{AG-JEPA} latent space, predicting future diagnosis and regional change directly from latent embeddings without synthesizing follow-up scans. To the best of our knowledge, this is the first approach to model AD progression as latent-space displacements within a frozen JEPA representation.

\item \textbf{Evaluation across diagnostic and progression tasks.} We evaluate our framework on \href{https://adni.loni.usc.edu}{ADNI} across five downstream tasks: CN versus MCI, CN versus AD, stable versus progressive MCI, cognitive score prediction and time-to-conversion forecasting. We further evaluate generalization on the \href{https://huggingface.co/datasets/Forithmus/MR-RATE}{MR-RATE}, \href{https://aibl.org.au}{AIBL} and \href{https://www.ucl.ac.uk/brain-sciences/ion/research/research-centres/dementia-research-centre/research-clinical-trials/minimal-interval-resonance-imaging-alzheimers-disease-miriad}{MIRIAD} datasets. \texttt{AG-JEPA} outperforms existing baselines across all diagnostic tasks, achieving its largest gain on the challenging CN versus MCI classification task with an improvement of 8.4 percentage points over the strongest baseline. It also improves CN versus AD classification by 7.1 percentage points over \texttt{Neuro-JEPA} in an out-of-distribution evaluation on MR-RATE. For progression modeling, \texttt{Prog-AG-JEPA} improves future-diagnosis AUROC by up to 1.7 points over the strongest generative baseline and achieves the lowest errors in regional volume change and average cognitive score change across three cognitive assessments.

\item \textbf{Representation analysis and interpretability.} We demonstrate that \texttt{AG-JEPA} learns anatomically organized and diagnostically discriminative representations through latent-space and attention-map analyses.
Its anatomical tokens provide direct per-region interpretability, linking learned representations to specific regions. 

\end{enumerate}

\section{Related Work}
\label{sec:related}

\textbf{Neuroimaging Foundation Models.} Brain-MRI foundation models such as \texttt{Neuro-JEPA}~\citep{huang2026neurojepa}, \texttt{NeuroVFM}~\citep{kondepudi2026neurovfm}, and \texttt{BrainIAC}~\citep{tak2026brainiac} learn representations at patch-grid locations using global self-supervised objectives. While these representations transfer well, they remain anchored to spatial positions rather than anatomically defined regions, which is limiting in Alzheimer's disease where the earliest pathological changes emerge in specific neuroanatomical structures before becoming more widespread~\citep{petersen2009adni}. Anatomy has been incorporated into pretraining in other ways, such as \texttt{VoCo}, which exploits the consistent spatial organization of anatomy by predicting the origin of sub-volume crops~\citep{wu2026voco}. In contrast, \texttt{AG-JEPA} explicitly models anatomically defined regions as latent tokens, supervises them with region-level objectives, and constrains their attention, yielding region-specific representations that remain interpretable at inference.

\textbf{Forecasting Disease Progression.} Most progression models forecast future neurodegeneration by generating a follow-up scan. \texttt{BrLP} conditions latent diffusion on subject metadata and auxiliary estimates of regional volume change~\citep{puglisi2025brain}, while \texttt{AG-LDM} jointly models baseline anatomy, follow-up anatomy, and clinical covariates in an end-to-end diffusion framework~\citep{wan2026agldm}. \texttt{Latent Drift} discretises inter-visit latent residuals using finite scalar quantisation and autoregressively generates drift tokens for follow-up MRI synthesis~\citep{feng2026latentdrift}. 
In contrast, \texttt{Prog-AG-JEPA} forecasts progression as conditioned latent displacements in the anatomically grounded \texttt{AG-JEPA} space, capturing diagnostic, cognitive, and regional structural changes without synthesizing follow-up images.
\section{Methods}
\label{sec:methods}

Our framework has two components. First, \texttt{AG-JEPA} learns to represent each brain MRI scan using patch tokens anchored to spatial locations and anatomical tokens anchored to named regions. Training couples these representations and guides each anatomical token to draw on the patches containing its region, supporting region-wise interpretation. Second, \texttt{Prog-AG-JEPA} builds on this representation and operates on the frozen encoder, predicting how a scan's patch and region embeddings will change by a follow-up visit, conditioned on the follow-up interval, current age, and regional volumes. Adding the predicted changes to the current embeddings yields future latent states without image synthesis. 

\begin{figure*}[!t]
\centering
\includegraphics[width=\textwidth]{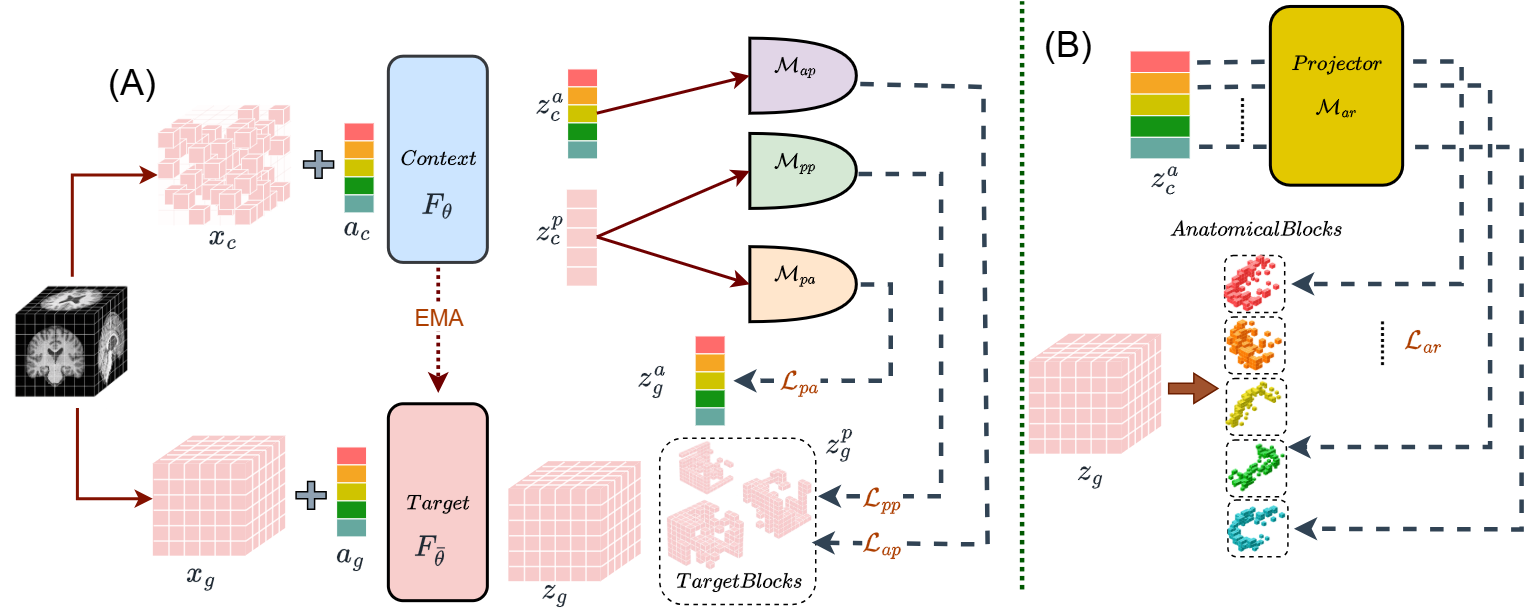}

\caption{Overview of \texttt{AG-JEPA}. \textbf{(A)}~The two encoder branches, each carrying one anatomical token per region, and the three prediction heads operating across the patch and anatomical token streams. \textbf{(B)}~The anatomy projector $\mathcal{M}_{\mathrm{ar}}$ used to align each anatomical token with its corresponding region representation. Section~\ref{sec:anatomy-objectives} gives the formulation.}

\label{fig:agjepa-architecture}
\vspace{-10pt}
\end{figure*}

\subsection{Joint Embedding Predictive Architecture} \label{sec:JEPA} 

We follow the masked feature-prediction pretraining framework of \texttt{V-JEPA 2}~\citep{assran2025vjepa2}, which predicts representations of masked input regions rather than reconstructing pixels~\citep{lecun2022path}. We adapt its video formulation to volumetric brain MRI using 3D spatial patches. Our backbone encoder follows the \texttt{Neuro-JEPA} architecture~\citep{huang2026neurojepa}, a 3D Vision Transformer with Mixture-of-Experts (MoE) blocks, which we adopt as the context and target encoders $\mathcal{F}_{\theta}$ and $\mathcal{F}_{\bar\theta}$ respectively. Given an input volume $x$, target blocks are masked and $x_c$ denotes the visible context patches, with $g$ indexing the target positions on the patch grid. The context encoder sees only the visible context patches and embeds them as $z^p_c=\mathcal{F}_{\theta}(x_c)$, while the target encoder encodes the complete volume, $z_g=\mathcal{F}_{\bar\theta}(x)$, from which the prediction targets $z^p_g$ are taken at the target positions. The target encoder is updated by an exponential moving average of the context encoder's weights, $\bar\theta \leftarrow m\,\bar\theta + (1-m)\,\theta$, rather than by backpropagation. The predictor $\mathcal{M}_{\mathrm{pp}}$ estimates the targets from the context embeddings and target positions $p_g$, giving $\hat{z}^p_g=\mathcal{M}_{\mathrm{pp}}(z^p_c,p_g)$. The context encoder and predictor are trained jointly by minimizing a smooth-$L_1$ loss $\ell$ between $\hat{z}^p_g$ and $z^p_g$.

\subsection{Neuroanatomical Regions} \label{sec:anatomy-tokenization} \paragraph{Anatomical Basis.} Alzheimer's disease produces some of its earliest structural changes in the medial temporal lobe, where the entorhinal cortex, hippocampus, and amygdala undergo progressive atrophy~\citep{braak1991staging,poulin2011amygdala}. This tissue loss is accompanied by enlargement of adjacent cerebrospinal fluid spaces, particularly the temporal horn and lateral ventricle~\citep{frisoni2002temporalhorn,apostolova2012ventricular}. We therefore focus on these five bilateral regions, denoted by $R$, which capture both direct markers of tissue loss and indirect markers of neurodegeneration through ventricular expansion. They are located with FastSurfer, an off-the-shelf pretrained labeling tool~\citep{henschel2020fastsurfer}, and its maps are used only during pretraining. No diagnostic or manually annotated label is used, and downstream probing and inference take the scan alone. 

\paragraph{Coverage Matrix.} Let $r\in\{1,\ldots,R\}$ index the regions and $n\in\{1,\ldots,N\}$ the patches. The map of region $r$ is average-pooled to the patch resolution and rescaled to sum to one. Collecting the rows gives a coverage matrix $C\in\mathbb{R}^{R\times N}$, where $C_{r,n}$ is the share of region $r$ within patch $n$; hence, each row is a distribution over patches. Each patch is weighted by the fraction of the region it contains rather than assigned exclusively to a single region, which is important for regions that are thin or small relative to a patch. These weights average the patch embeddings into one representation per region, \begin{linenomath} \begin{equation} \label{eq:pooled} z^{\mathrm{reg}}_{g,r} = \sum\nolimits_{n=1}^{N} C_{r,n}\,z_{g,n}, \end{equation} \end{linenomath} where $z_{g,n}$ is the embedding of $n^{th}$ patch of the full target embedding $z_g$ from the target encoder. Appendix~\ref{app:coverage} elaborates further.


\paragraph{Context Sampling.} Unconstrained block masking can hide a small region entirely, leaving its token without visible evidence. We therefore use anatomical region maps to guide which patches remain visible to the context encoder. Specifically, we identify patches substantially covered by the anatomical regions and assign a quarter of them to the context. Every region therefore retains visible context, while most of its patches remain masked and their embeddings must be predicted.

\subsection{AG-JEPA Architecture} \label{sec:anatomy-objectives} 
\texttt{AG-JEPA} extends this masked-prediction framework with anatomy-aware latent tokens that explicitly represent clinically relevant brain structures. 
For each anatomical region, we introduce a learnable anatomy token, denoted by $a_c$ and $a_g$ in the context and target branches, respectively. These tokens are prepended to the patch sequence and jointly processed with the patch tokens by the encoder, producing anatomical embeddings $z^a_c$ and $z^a_g$ alongside the patch embeddings $z^p_c$ and $z^p_g$. The anatomy tokens are intended to summarize the structural state of their corresponding regions while interacting with the patch representation through self-attention. The resulting context and target embeddings are given by $(z^a_c,z^p_c)=\mathcal{F}_{\theta}(x_c,a_c)$ and $(z^a_g,z_g)=\mathcal{F}_{\bar{\theta}}(x,a_g)$, with target patch embeddings $z_g^p$ selected from $z_g$ at the masked positions. 
As illustrated in Figure~\ref{fig:agjepa-architecture}, \texttt{AG-JEPA} augments the standard JEPA framework with three prediction heads and an anatomy projector operating over the patch and anatomical token streams, together with attention regularization that guides each token toward its corresponding anatomical region. Each prediction head is a lightweight Vision Transformer, while the anatomy projector is a multilayer perceptron (MLP). More details are in Appendix~\ref{app:agjepa-implementation}.

\paragraph{Anatomy-Guided Predictive Objectives.} To learn anatomy-aware representations, \texttt{AG-JEPA} employs three complementary predictive objectives and a region-alignment objective, together with attention regularization that guides the anatomical tokens toward their corresponding regions. For objectives defined at the region level, losses are averaged across regions. 
Every target is layer-normalized and carries a stop-gradient, so no gradient reaches target encoder.


\paragraph{Patch Prediction.} The patch predictor $\mathcal{M}_{\mathrm{pp}}$ of Section~\ref{sec:JEPA} predicts masked patch embeddings from visible context embeddings 
$\mathcal{L}_{\mathrm{pp}}=\ell\big(\mathcal{M}_{\mathrm{pp}}(z^p_c,p_g),z^p_g\big)$. This term holds the patch representation to the volume at full spatial resolution, so that the anatomical summary is formed from detailed features. 

\paragraph{Patch Prediction from Anatomy Tokens.} The anatomy-guided patch predictor $\mathcal{M}_{\mathrm{ap}}$ predicts the same masked patch embeddings as $\mathcal{M}_{\mathrm{pp}}$, but receives only anatomical embeddings and target positions, without direct access to the context patch embeddings: $\mathcal{L}_{\mathrm{ap}}=\ell\big(\mathcal{M}_{\mathrm{ap}}(z^a_c,p_g),z^p_g\big)$. The tokens are thus the only route by which information about the scan can reach this predictor, compressing information that would otherwise be spread across the patch embeddings.

\paragraph{Anatomy Tokens Prediction.} The anatomy predictor $\mathcal{M}_{\mathrm{pa}}$ runs in the opposite direction, recovering the target tokens from the visible patches, $\mathcal{L}_{\mathrm{pa}}=\ell\big(\mathcal{M}_{\mathrm{pa}}(z^p_c),z^a_g\big)$. The asymmetry gives the term its content: $z^a_g$ is produced from the complete volume, whereas $z^p_c$ covers only the visible quarter, so a whole region must be inferred from partial evidence. Together with $\mathcal{L}_{\mathrm{ap}}$, this objective encourages the patch and anatomical streams to describe the same volume consistently at two levels of detail.

\paragraph{Anatomy Projection.} The anatomy projector $\mathcal{M}_{\mathrm{ar}}$ maps each context anatomical embedding onto the pooled representation of its region from Eq.~\eqref{eq:pooled}, scored one region at a time: \begin{linenomath} \begin{equation} \label{eq:lar} \mathcal{L}_{\mathrm{ar}} = \operatorname*{mean}_{r\in{R}} \ell\!\left( \mathcal{M}_{\mathrm{ar}}(z^a_{c,r}), z^{\mathrm{reg}}_{g,r} \right). \end{equation} \end{linenomath} This term aligns each token's content with its corresponding region, complementing the attention localization introduced below. 
The projector is a single multilayer perceptron shared across regions, since per-region heads could absorb the distinction between regions themselves, whereas one map for all of them requires that difference to sit in the tokens.

\begin{figure}[!t]
\centering
\includegraphics[width=\columnwidth]{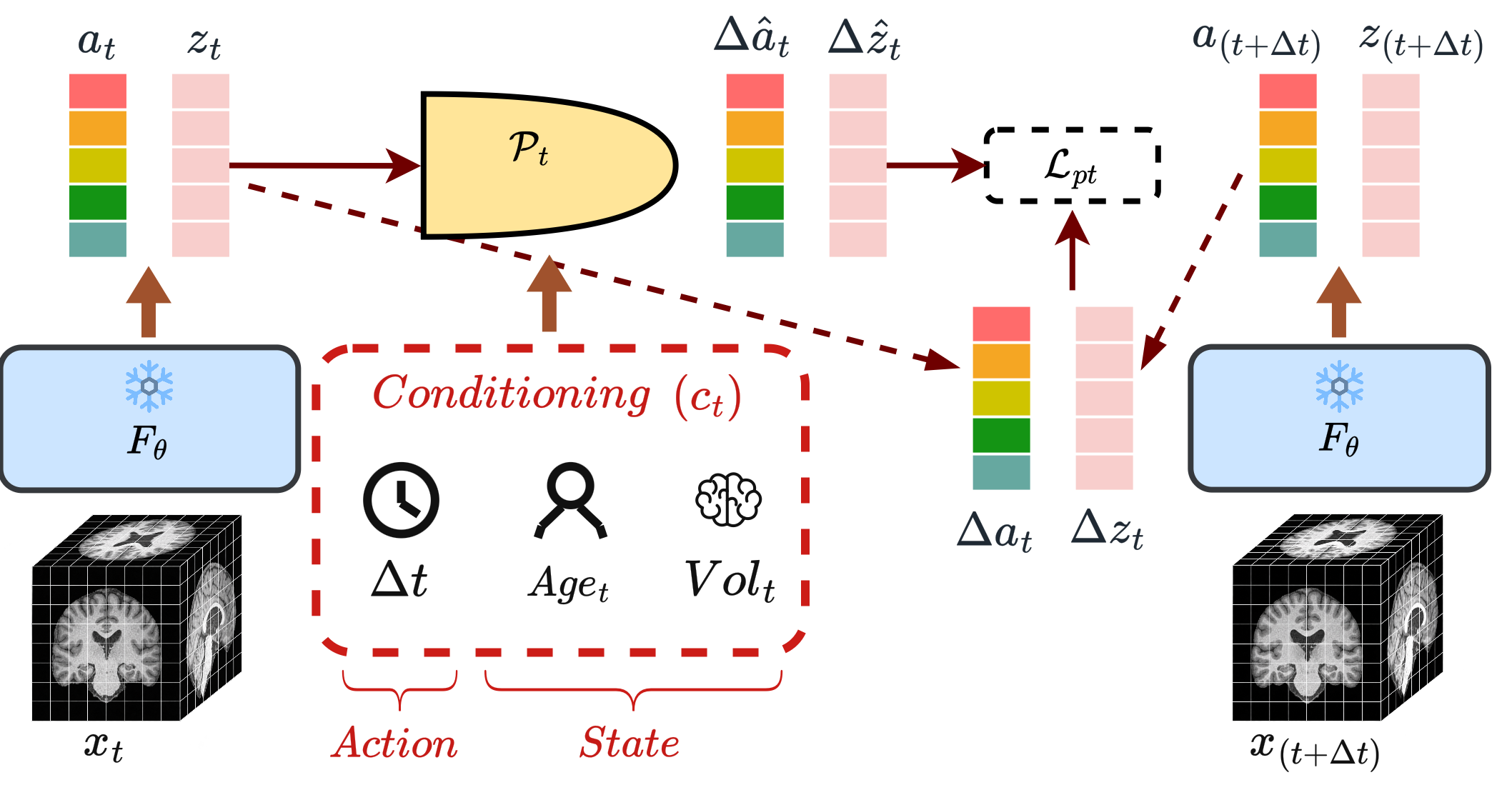}
\caption{\texttt{Prog-AG-JEPA}. The progression predictor $\mathcal{P}_t$ uses the current representations $(z_t,a_t)$ and conditioning $c_t=(\Delta t,\mathrm{Age}_t,\mathrm{Vol}_t)$ to predict the future latent and anatomical drifts $(\Delta\hat{z}_t,\Delta\hat{a}_t)$, trained against the corresponding observed drifts $(\Delta z_t,\Delta a_t)$.}

\label{fig:prog-agjepa}
\end{figure}
\begin{table*}[!t]
\floatconts
{tab:adni-main}
{\caption{Quantification of \texttt{AG-JEPA} features on the ADNI Test set across downstream tasks. }}

\resizebox{\textwidth}{!}{%
  \begin{tabular}{lcccccccc}
  \toprule
  & \multicolumn{3}{c}{Diagnosis Classification (AUROC)} & \multicolumn{2}{c}{C-Index\,/\,MAE} & \multicolumn{3}{c}{Cognitive Scores (MAE)}\\
  \cmidrule(lr){2-4}\cmidrule(lr){5-6}\cmidrule(lr){7-9}
  Method & CN/MCI\,$\uparrow$ & CN/AD\,$\uparrow$ & sMCI/pMCI\,$\uparrow$ & TTC\,$\uparrow$ & Age\,$\downarrow$ & MMSE\,$\downarrow$ & ADAS-13\,$\downarrow$ & CDR-SB\,$\downarrow$\\
  \midrule
  \texttt{NeuroVFM} & $0.717{\pm}0.050$ & $0.857{\pm}0.050$ & $0.577{\pm}0.071$ & $0.559{\pm}0.059$ & $5.71{\pm}0.20$ & $2.716{\pm}0.151$ & $7.707{\pm}0.416$ & $1.623{\pm}0.093$\\
  \texttt{Neuro-JEPA} & $0.720{\pm}0.050$ & $0.876{\pm}0.040$ & $0.628{\pm}0.075$ & $0.556{\pm}0.064$ & $5.70{\pm}0.20$ & $2.503{\pm}0.147$ & $7.365{\pm}0.419$ & $1.495{\pm}0.091$\\
  \rowcolor{agrow}
  \texttt{AG-JEPA} & $\mathbf{0.804{\pm}0.042}$ & $\mathbf{0.887{\pm}0.042}$ & $\mathbf{0.689{\pm}0.073}$ & $\mathbf{0.612{\pm}0.053}$ & $\mathbf{5.63{\pm}0.15}$ & $\mathbf{2.460{\pm}0.153}$ & $\mathbf{7.068{\pm}0.386}$ & $\mathbf{1.473{\pm}0.086}$\\
\bottomrule
\end{tabular}}
\end{table*}

\paragraph{Supervising Where the Tokens Attend.} These four terms constrain what a token contains, but not where it draws that content from: a token could reproduce its region's content while attending across the whole brain. Let $\phi_r$ be the distribution over context patches that token $r$ attends to in the final encoder block, averaged over heads, and $t_r$ the $r$-th row of $C$ restricted to those patches and renormalized, 
%
\begin{linenomath} \begin{equation} \label{eq:attn} \begin{aligned} \mathcal{L}_{\mathrm{attn}}={}&-\lambda_{\mathrm{attn}}\operatorname*{mean}_{r\in{R}} \left\langle t_r,\log\phi_r\right\rangle \\ &+\lambda_{\mathrm{ov}}\operatorname*{mean}_{r\neq s,\;r,s\in{R}} \left\langle\phi_r,\phi_s\right\rangle. \end{aligned} \end{equation} \end{linenomath}
The first term pulls each token's attention onto its own region, minimized at $\phi_r=t_r$; the second keeps the maps distinct by penalizing tokens that attend to the same patches. Since coverage targets overlap where regions adjoin, $\lambda_{\mathrm{ov}}$ stays small relative to $\lambda_{\mathrm{attn}}$.

\paragraph{Full objective.} The objectives are combined as $\mathcal{L}=\mathcal{L}_{\mathrm{pp}}+\mathcal{L}_{\mathrm{ap}}+\mathcal{L}_{\mathrm{pa}}+\lambda_{\mathrm{ar}}\mathcal{L}_{\mathrm{ar}}+\mathcal{L}_{\mathrm{attn}}$, with weights and schedules given in the implementation details. The coverage matrix $C$ enters at three points only, through the pooled targets of $\mathcal{L}_{\mathrm{ar}}$, the attention targets $t_r$, and the context sampling; no other anatomical information is used during training, and none at all afterwards. Table~\ref{tab:heads} in Appendix~\ref{app:components} summarizes the prediction components, and Table~\ref{tab:ablation} reports the contribution of each component.

  

\subsection{\texttt{Prog-AG-JEPA}: Latent Progression Modeling}
\label{sec:prog}

\texttt{Prog-AG-JEPA} is a lightweight head, trained on top of the frozen encoder, that jointly predicts changes in the anatomical and patch representations between two visits. Figure~\ref{fig:prog-agjepa} shows the head, the two visits it compares, and the variables it conditions on. Let $x_t$ and $x_{(t+\Delta t)}$ be two scans of the same subject, and let $(a_t,z_t)$ and $(a_{(t+\Delta t)},z_{(t+\Delta t)})$ denote their anatomical-token and patch-token representations. The head $\mathcal{P}_t$ predicts both drifts from the current representations and three conditioning variables, $(\Delta\hat{a}_t,\Delta\hat{z}_t)=\mathcal{P}_t(a_t,z_t,c_t)$ with $c_t=(\Delta t,\mathrm{Age}_t,\mathrm{Vol}_t)$, where $\Delta t$ is the interval between scans, $\mathrm{Age}_t$ is the subject's current age, and $\mathrm{Vol}_t$ holds the $R$ regional volumes of the current scan. The observed drifts are $\Delta a_t=a_{(t+\Delta t)}-a_t$ and $\Delta z_t=z_{(t+\Delta t)}-z_t$, yielding residual predictions $\hat{a}_{(t+\Delta t)}=a_t+\Delta\hat{a}_t$ and $\hat{z}_{(t+\Delta t)}=z_t+\Delta\hat{z}_t$. The head is trained with a joint smooth-$L_1$ objective, $\mathcal{L}_{\mathrm{pt}}=\lambda_a\ell(\Delta\hat{a}_t,\Delta a_t)+\lambda_z\ell(\Delta\hat{z}_t,\Delta z_t)$, after standardizing predicted and observed drifts using training-pair statistics. Progression is therefore modeled as a conditioned step within a fixed, region-organized representation in the latent space, rather than synthesizing and encoding the follow-up image.

\begin{figure}
\centering
\includegraphics[width=0.9\linewidth]{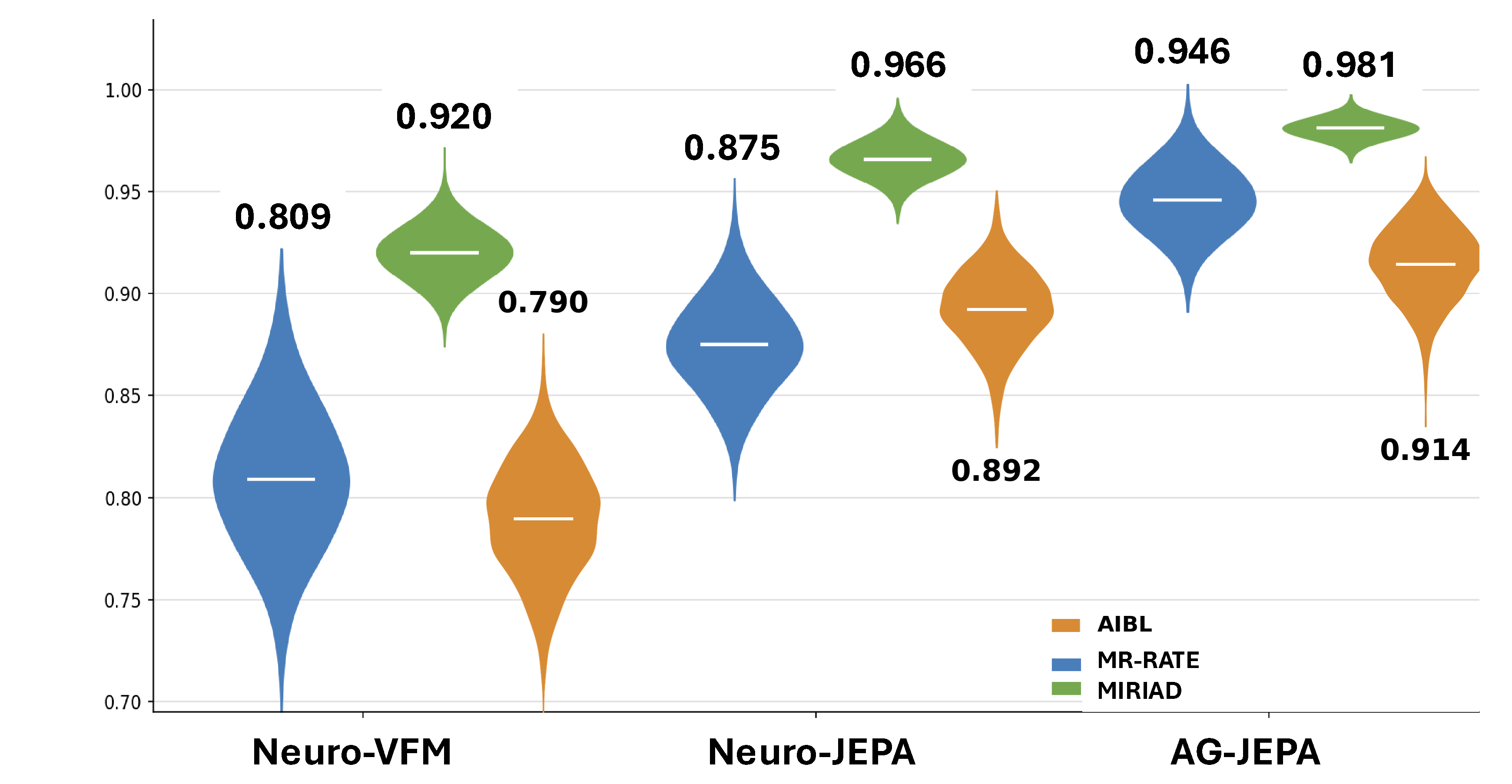}
\caption{CN/AD AUROC on cross-cohort MR-RATE, MIRIAD and AIBL, with probes fitted on ADNI train.
}
\label{fig:mrrate-cnad}
\end{figure}

\begin{table*}[!t]
\floatconts
  {tab:prog-cog}
  {\caption{\texttt{Prog-AG-JEPA}'s predicted future latents across downstream tasks on ADNI Test.}}
  {
  \setlength{\tabcolsep}{3pt}
  \resizebox{1\textwidth}{!}{%
  \begin{tabular}{lccccccccc}
  \toprule
  & \multicolumn{2}{c}{Diagnosis (AUROC)}
  & \multicolumn{3}{c}{Cognitive Score Change (MAE)}
  & \multicolumn{4}{c}{Volume Change (MAE)} \\
  \cmidrule(lr){2-3}
  \cmidrule(lr){4-6}
  \cmidrule(lr){7-10}
  Method
  & CN/AD $\uparrow$
  & CN/MCI $\uparrow$
  & MMSE $\downarrow$
  & ADAS-13 $\downarrow$
  & CDR-SB $\downarrow$
  & Hipp. $\downarrow$
  & Amyg. $\downarrow$
  & Temp. Horn $\downarrow$
  & Lat. Vent. $\downarrow$ \\
  \midrule

  \texttt{Latent Drift}
  & $0.719{\pm}0.053$
  & $0.696{\pm}0.058$
  & $2.010{\pm}0.174$
  & $4.752{\pm}0.245$
  & $1.083{\pm}0.058$ 
  & $0.024{\pm}0.001$
  & $0.016{\pm}0.001$
  & $0.035{\pm}0.004$
  & $0.235{\pm}0.015$ \\

  \texttt{BrLP}
  & $0.827{\pm}0.060$
  & $0.751{\pm}0.065$
  & $2.050{\pm}0.165$
  & $4.550{\pm}0.211$
  & $1.030{\pm}0.065$
  & $0.022{\pm}0.002$
  & $0.013{\pm}0.001$
  & $0.030{\pm}0.003$
  & $0.215{\pm}0.018$ \\

  \texttt{AG-LDM}
  & $0.818{\pm}0.060$
  & $0.741{\pm}0.067$
  & $\mathbf{1.889{\pm}0.184}$
  & $4.386{\pm}0.229$
  & $0.929{\pm}0.073$
  & $0.020{\pm}0.001$
  & $0.012{\pm}0.001$
  & $0.032{\pm}0.004$
  & $0.243{\pm}0.017$ \\

  \rowcolor{agrow}
  \texttt{Prog-AG-JEPA}
  & $\mathbf{0.844{\pm}0.052}$
  & $\mathbf{0.765{\pm}0.060}$
  & $1.893{\pm}0.159$
  & $\mathbf{4.246{\pm}0.230}$
  & $\mathbf{0.911{\pm}0.064}$
  & $\mathbf{0.018{\pm}0.001}$
  & $\mathbf{0.010{\pm}0.001}$
  & $\mathbf{0.024{\pm}0.003}$
  & $\mathbf{0.185{\pm}0.013}$ \\

  \bottomrule
  \end{tabular}}
  }
\end{table*}

\section{Experimental Details}
\label{sec:exp}
 
We evaluate the representations learned by \texttt{AG-JEPA} on five tasks spanning diagnosis and progression: CN/MCI, CN/AD, stable/progressive MCI, time-to-conversion (MCI to AD) prediction, and cognitive score prediction for three assessments with their score range: MMSE (0--30), ADAS-13 (0--85), and CDR-SB (0--18). In addition, we evaluate age prediction. The encoder is frozen throughout and assessed via linear probing, ensuring that the evaluation reflects the quality of the learned representations rather than task-specific adaptation. 
We further assess whether the future time-point latents predicted by \texttt{Prog-AG-JEPA} preserve progression-relevant information through regional-volume and cognitive-score change estimation between two visits, as well as diagnostic classification.

\textbf{Dataset Description.} We use T1-weighted CN, MCI, and AD MRI from ADNI~\citep{petersen2009adni}, MR-RATE~\citep{mrrate2026}, AIBL~\citep{ellis2009aibl}, and MIRIAD~\citep{malone2013miriad}. Subject-disjoint ADNI train supports model development, ADNI Test provides within-cohort evaluation, and MR-RATE, AIBL and MIRIAD assess cross-cohort transfer without adaptation. Pre-processing and task-wise cohort composition details are in Appendix~\ref{app:data-details}.


\textbf{Baselines.} For representation learning, we select \texttt{Neuro-JEPA} and \texttt{NeuroVFM}, neuroimaging JEPA-based foundation models that have been shown to effectively learn representations for discriminating CN/AD~\citep{huang2026neurojepa,kondepudi2026neurovfm}. For progression, we select \texttt{BrLP}, \texttt{AG-LDM} and \texttt{Latent Drift}, which model longitudinal disease trajectories by generating the follow-up images~\citep{puglisi2025brain,wan2026agldm,feng2026latentdrift}. All models were trained using their official repositories.

\textbf{Implementation Details.} \texttt{AG-JEPA} is pretrained for 300 epochs using AdamW with a constant learning rate of $3\times10^{-5}$, global batch size 64, and an EMA target encoder (momentum $0.99925$). Foreground-aware masking uses a target ratio of $75\%$ while ensuring that one quarter of regional patches remain in the context. \texttt{Prog-AG-JEPA} is trained for 30 epochs with a learning rate of $3\times10^{-4}$, and classification probes employ balanced logistic regression.
Full implementation details are provided in Appendix~\ref{app:implementation}.


\vspace{-5pt}
\section{Results and Discussion}
\label{sec:results}

\textbf{Diagnosis Tasks on ADNI.} Table~\ref{tab:adni-main} reports results on the ADNI Test set. \texttt{AG-JEPA} outperforms both baselines across all five tasks, improving over the strongest by $8.4$ percentage points on CN/MCI, $6.1$ percentage points on sMCI/pMCI, and $5.3$ points on time-to-conversion prediction. The largest gains are observed on tasks where disease- and progression-related signals are more subtle, while consistent improvements are maintained on CN/AD classification, age prediction, and cognitive-score regression.



\textbf{Cross-Cohort Generalization.} Figure~\ref{fig:mrrate-cnad} evaluates ADNI-trained CN/AD probes on MR-RATE, MIRIAD and AIBL without adaptation. \texttt{AG-JEPA} improves over the strongest baseline, \texttt{Neuro-JEPA}, by $7.1$, $1.5$ and $2.2$ percentage points on MR-RATE, MIRIAD and AIBL, respectively. Appendix~\ref{app:representation-generalization} further shows clearer cross-cohort diagnostic separation, with less cohort-dependent clustering and class overlap, as compared to the baselines.

\textbf{Future Latent Prediction.} Table~\ref{tab:prog-cog} presents results for future diagnosis and cognitive-score change prediction on longitudinal scan pairs. \texttt{Prog-AG-JEPA} is trained on ADNI Train and evaluated on ADNI Test. For future diagnosis, we achieve the highest AUROC across both tasks, outperforming \texttt{BrLP} by $1.7$ and $1.4$ percentage points for CN/AD and CN/MCI, respectively. For cognitive-score change, we achieve the lowest MAE on ADAS-13 and CDR-SB and remain comparable to the best baseline on MMSE. These results show that the predicted latents retain progression-relevant diagnostic and cognitive information.
\begin{figure}[!b]
\centering
\includegraphics[width=\columnwidth]{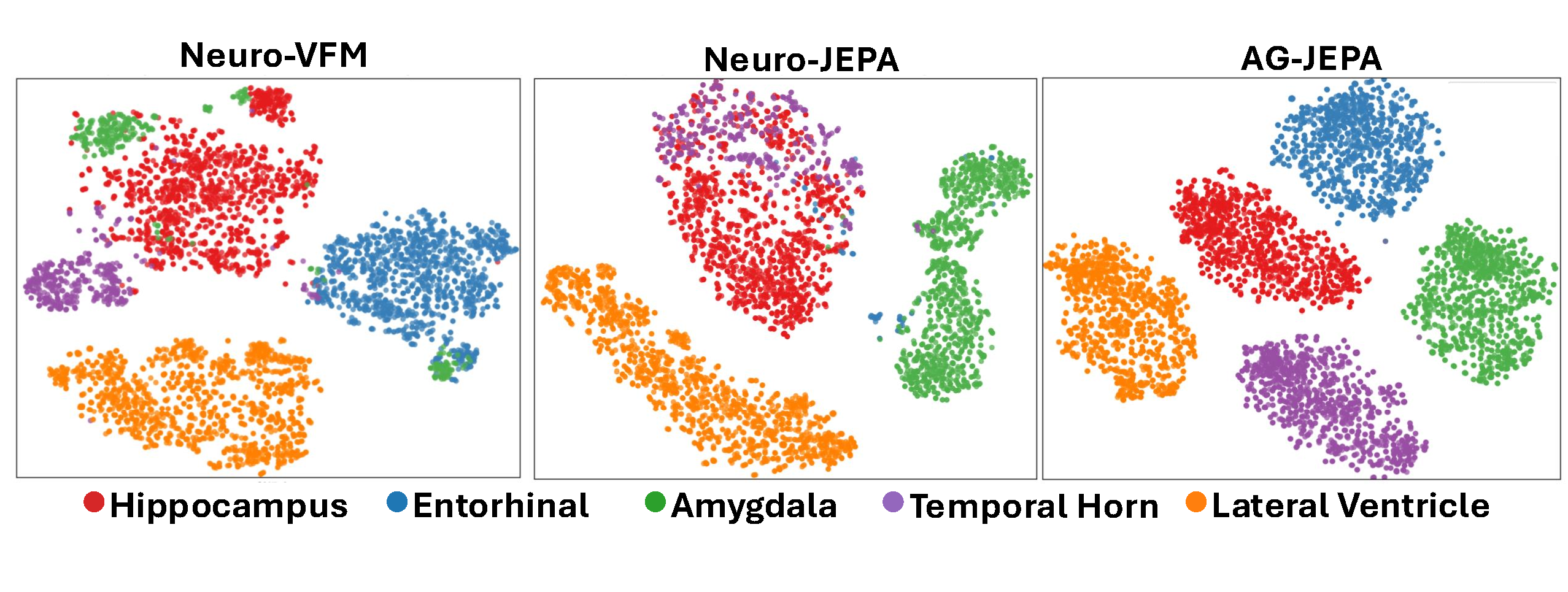}
\caption{t-SNE of region-specific representations on ADNI Test. \texttt{AG-JEPA} anatomy-token embeddings form compact, separable regional clusters, while baseline regional patch embeddings are more overlapping.}
\label{fig:qual-region-emb}
\end{figure}


\textbf{Regional Volume Change Estimation.} The volume-change results in Table~\ref{tab:prog-cog} quantify regional volume-change errors (with metrics and longitudinal controls in Appendix~\ref{app:regional-volume-eval}), while Figure~\ref{fig:qual-vol} visualizes the corresponding signed changes within the regional masks. \texttt{Prog-AG-JEPA} achieves the lowest MAE for every evaluated region and most closely reproduces the ground-truth patterns of atrophy and expansion. Thus, despite operating entirely in latent space, it localizes region-specific structural changes without requiring future-image generation or explicit diagnostic-label conditioning.

All reported statistical significance analyses for the quantitative evaluations are provided in Appendix~\ref{app:paired-validation}.

\begin{figure}[!t]
\centering
\includegraphics[width=1\columnwidth]{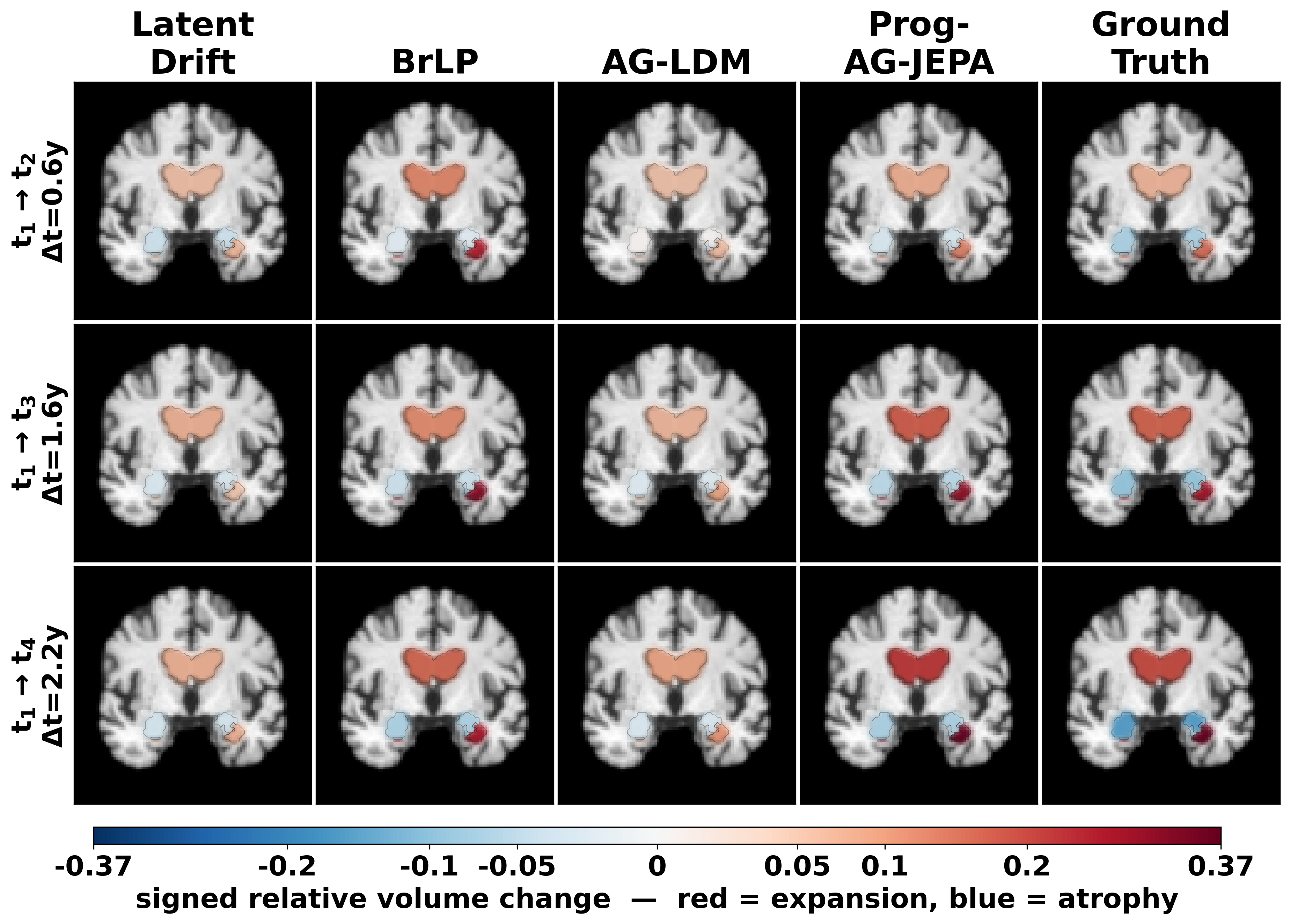}
\caption{Region-wise signed relative volume change across three longitudinal intervals,
shown within the corresponding anatomical masks for the ground truth and each model.
Blue indicates atrophy (decreased volume), while red indicates expansion (increased
volume).}
\label{fig:qual-vol}
\end{figure}

\begin{figure}
\centering
\includegraphics[width=\columnwidth]{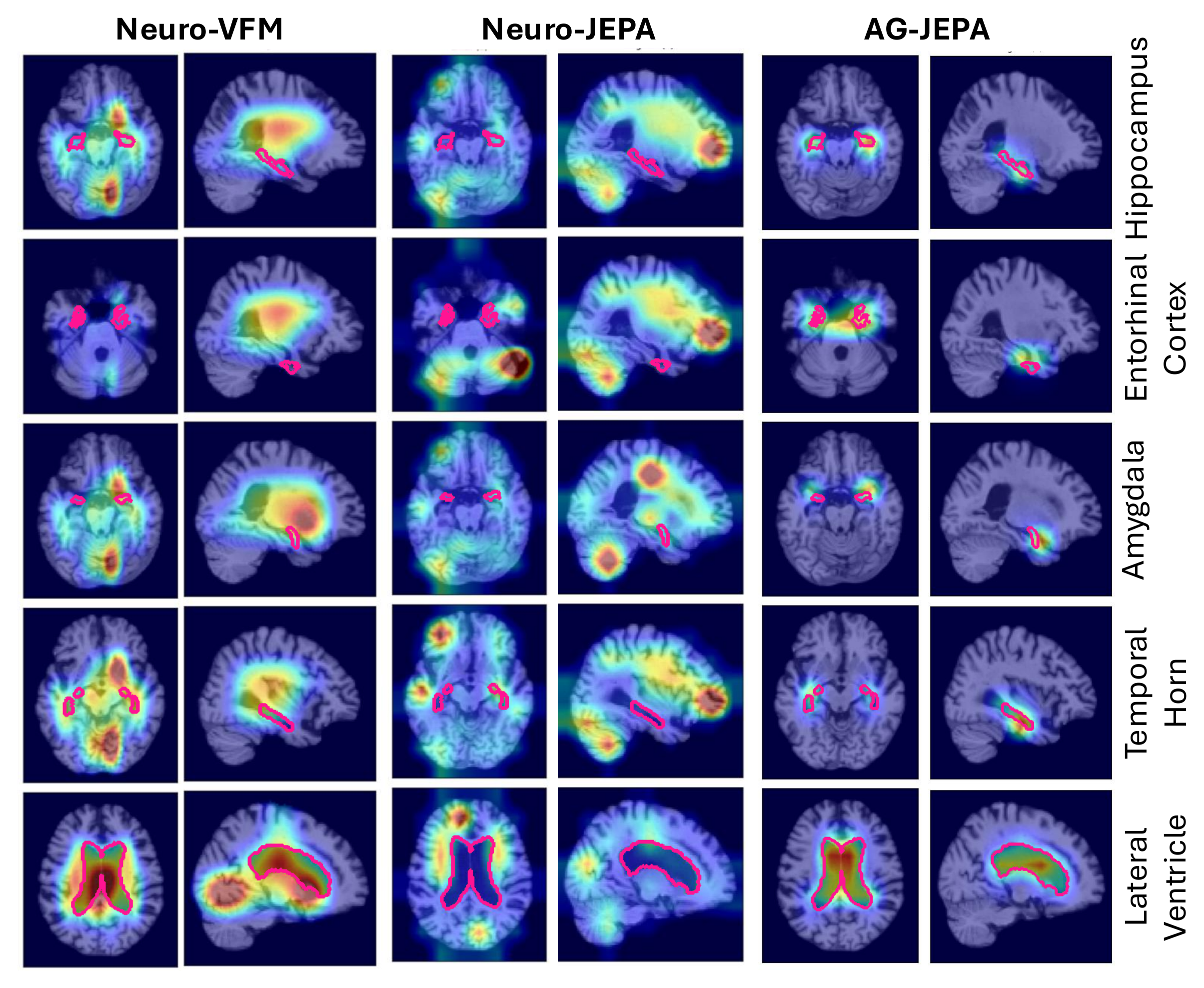}
\caption{Axial and sagittal attention maps for an MCI subject across five anatomical regions, with anatomical segmentation boundaries shown in pink. \texttt{AG-JEPA} anatomy-token attention remains region-localized, whereas baseline attention maps are diffuse.
Baselines show one global encoder attention-rollout map across region views.
}
\label{fig:qual-region}
\end{figure}


\textbf{Attention and Representational Analysis.} Figure~\ref{fig:qual-region} assesses where each anatomical token gathers its content. Across both views, each \texttt{AG-JEPA} attention map $\phi_r$ localizes within the corresponding labeled region, whereas the baseline maps remain diffuse and largely similar across regions. Figure~\ref{fig:qual-region-emb} further shows that \texttt{AG-JEPA} anatomy-token embeddings form compact, separable regional clusters, while baseline regional patch embeddings are overlapping. This shows that our introduced objectives ($\mathcal{L}_{\mathrm{ar}}$ and $\mathcal{L}_{\mathrm{attn}}$) anatomically organize the representational space. 
Further, disease information is assessed separately through token-level probes (Table~\ref{tab:revision-token-probes}) and cross-cohort diagnostic visualization (Appendix~\ref{app:representation-generalization}).

Overall, these results demonstrate that \texttt{AG-JEPA} learns anatomically organized representations that are more discriminative, and robust across cohorts, hence better suited for longitudinal disease modeling.

\begin{table}[t]
\centering
\caption{Ablation of \texttt{AG-JEPA} objectives. Probes fitted on ADNI train, tested on ADNI Test.}
\label{tab:ablation}
\footnotesize\setlength{\tabcolsep}{1.9pt}
\begin{tabular}{@{}ccccc@{\hspace{5pt}}ccc@{}}
\toprule
\multicolumn{5}{c@{\hspace{5pt}}}{Objective terms} & \multicolumn{3}{c}{AUROC $\uparrow$}\\
\cmidrule(r){1-5}\cmidrule(l){6-8}
$\mathcal{L}_{\mathrm{pp}}$ & $\mathcal{L}_{\mathrm{ap}}$ & $\mathcal{L}_{\mathrm{pa}}$ & $\mathcal{L}_{\mathrm{ar}}$ & $\mathcal{L}_{\mathrm{attn}}$ &
\shortstack{CN/\\MCI} & \shortstack{CN/\\AD} & \shortstack{sMCI/\\pMCI}\\
\midrule
$\bullet$ &           &           &           &           & $0.73{\pm}0.05$ & $0.87{\pm}0.04$ & $0.55{\pm}0.08$\\
$\bullet$ &           & $\bullet$ &           &           & $0.69{\pm}0.05$ & $0.88{\pm}0.05$ & $0.58{\pm}0.08$\\
$\bullet$ & $\bullet$ &           &           &           & $0.75{\pm}0.05$ & $0.87{\pm}0.05$ & $0.55{\pm}0.08$\\
$\bullet$ & $\bullet$ & $\bullet$ &           & $\bullet$ & $0.79{\pm}0.05$ & $0.87{\pm}0.05$ & $0.66{\pm}0.08$\\
$\bullet$ & $\bullet$ & $\bullet$ & $\bullet$ &           & $0.78{\pm}0.04$ & $\mathbf{0.90{\pm}0.04}$ & $0.65{\pm}0.07$\\
\rowcolor{agrow}
$\bullet$ & $\bullet$ & $\bullet$ & $\bullet$ & $\bullet$ & $\mathbf{0.80{\pm}0.04}$ & $0.89{\pm}0.04$ & $\mathbf{0.69{\pm}0.07}$\\
\bottomrule
\end{tabular}
\end{table}

\textbf{Ablations.}To assess the contributions of different predictive objectives, anatomical region selection during training, and the learned anatomical tokens used at inference, we conduct three complementary ablations of \texttt{AG-JEPA}.

Table~\ref{tab:ablation} evaluates the role of anatomy-guided \textbf{predictive objectives} across disease-detection tasks. Adding anatomy-guided patch prediction ($\mathcal{L}_{\mathrm{ap}}$) provides a modest improvement over patch prediction alone ($\mathcal{L}_{\mathrm{pp}}$), indicating that the anatomy tokens begin to capture information relevant to the underlying anatomy. In contrast, anatomy-token prediction ($\mathcal{L}_{\mathrm{pa}}$) is less effective on its own, suggesting that patch and anatomy representations must first become informative before they can effectively supervise one another. Larger gains emerge when anatomy tokens are explicitly grounded to image regions through attention alignment ($\mathcal{L}_{\mathrm{attn}}$) and assigned region-specific identities through anatomy projection ($\mathcal{L}_{\mathrm{ar}}$). The full objective achieves strongest results overall,
that suggest that informative anatomy tokens ($\mathcal{L}_{\mathrm{ap}}$), alignment between anatomy and patch representations ($\mathcal{L}_{\mathrm{pa}}$), anatomical grounding ($\mathcal{L}_{\mathrm{attn}}$), and region identity ($\mathcal{L}_{\mathrm{ar}}$) provide complementary supervision and are most effective when learned jointly.
Additionally, to distinguish the need for anatomical representations from spatial awareness alone, we compare with a JEPA baseline that encourages similar embeddings for spatially overlapping sub-volumes (Appendix~\ref{app:anatomical-controls}).

\begingroup

Beyond the predictive objectives, Table~\ref{tab:revision-region-set} evaluates \textbf{anatomical region selection} during pretraining. Expanding from 5 to 18 regions has little effect on CN/AD discrimination, whereas CN/MCI performance drops in both ADNI and AIBL. This suggests that focusing on a precise set of AD-relevant regions is relatively better for distinguishing early-stage disease than broadening anatomical coverage.

\begin{table}[!ht]

\floatconts
{tab:revision-region-set}
{\caption{Five versus 18 anatomical regions (AUROC), with ADNI-fitted probes.}}
{\scriptsize\setlength{\tabcolsep}{2.5pt}
\begin{tabular}{@{}llcc@{}}
\toprule
Dataset & Task & 5 Regions & 18 Regions \\
\midrule
ADNI & CN/AD & 0.887 & 0.889 \\
ADNI & CN/MCI & 0.804 & 0.788 \\
AIBL & CN/AD & 0.905 & 0.900 \\
AIBL & CN/MCI & 0.705 & 0.690 \\
\bottomrule
\end{tabular}}
\end{table}

Further, to assess the disease information captured by the learned anatomical tokens, Table~\ref{tab:revision-token-probes} assesses \textbf{anatomical token contribution} with the pretrained encoder fixed, using single-token and leave-one-token-out linear probes. The hippocampal token is the strongest individual token and its removal gives the largest AUROC decrease across both tasks and cohorts. This supports disease relevance and partially complementary information within the tokens.

Additional \texttt{Prog-AG-JEPA} ablations include controls that assess future diagnosis and clinical change prediction using conditioning variables only, as well as conditioning variables combined with current imaging features (without predicting latent drift), to establish the need for latent drift modeling.(Appendices~\ref{app:conditioning-ablation}, \ref{app:token-validation}, and \ref{app:current-state-ablation}, respectively).  

\begin{table}[!ht]

\floatconts
{tab:revision-token-probes}
{\caption{Token-level diagnosis AUROC. Probes are fitted on ADNI Train; AIBL is unseen. Token-selection comparisons use the all-five-anatomical-token row as reference.}}
{\scriptsize\setlength{\tabcolsep}{2.5pt}
\begin{tabular}{@{}>{\raggedright\arraybackslash}p{0.38\columnwidth}cccc@{}}
\toprule
& \multicolumn{2}{c}{ADNI} & \multicolumn{2}{c}{AIBL} \\
\cmidrule(lr){2-3}\cmidrule(lr){4-5}
Tokens & CN/AD & CN/MCI & CN/AD & CN/MCI \\
\midrule
All anatomical + patch tokens & 0.887 & 0.804 & 0.905 & 0.705 \\
All five anatomical tokens & 0.885 & 0.801 & 0.908 & 0.703 \\
\midrule
\multicolumn{5}{@{}l}{Single token} \\
Hippocampus & 0.882 & 0.802 & 0.903 & 0.701 \\
Temporal horn & 0.880 & 0.768 & 0.875 & 0.698 \\
Entorhinal cortex & 0.870 & 0.742 & 0.894 & 0.694 \\
Amygdala & 0.854 & 0.708 & 0.859 & 0.684 \\
Lateral ventricle & 0.871 & 0.782 & 0.891 & 0.620 \\
\midrule
\multicolumn{5}{@{}l}{Leave one token out} \\
Hippocampus & 0.852 & 0.775 & 0.880 & 0.686 \\
Temporal horn & 0.869 & 0.782 & 0.896 & 0.690 \\
Entorhinal cortex & 0.871 & 0.789 & 0.896 & 0.691 \\
Amygdala & 0.880 & 0.795 & 0.897 & 0.701 \\
Lateral ventricle & 0.875 & 0.795 & 0.897 & 0.701 \\
\bottomrule
\end{tabular}}
\end{table}

\endgroup

\section{Conclusion}
\label{sec:conclusion}
We presented \texttt{AG-JEPA}, a JEPA framework that incorporates anatomical attribution into pretraining by learning region-specific tokens anchored to clinically relevant neuroanatomical structures. This anatomical prior yields more informative representations, improving diagnosis and progression prediction, particularly in the earliest stages of disease. We further introduced \texttt{Prog-AG-JEPA}, which models disease progression as trajectories in the learned representation space without synthesizing future scans. Beyond improved predictive performance, the learned anatomy tokens provide direct region-level interpretability from the encoder itself. Although instantiated using five neurodegeneration-relevant brain regions, the proposed framework is not tied to a fixed anatomical partition. Future work will explore richer and potentially learned anatomical priors, including whole-brain parcellations, enabling anatomically grounded brain foundation models that capture the spatial organization and temporal evolution of the brain. We believe such models can unify representation learning for diagnosis, prognosis, progression modeling, and disease understanding.


\bibliography{ref}

\clearpage
\onecolumn
\appendix
\section{Additional Architectural Details}
\label{app:architectural-details}

Section~\ref{sec:methods} describes the anatomical tokens and the objectives that train them. This appendix supplies the construction behind them: how the coverage matrix is computed from a segmentation, and what each prediction component contributes.

\subsection{Constructing the Coverage Matrix}
\label{app:coverage}

Section~\ref{sec:anatomy-tokenization} defines the coverage matrix; here we define its normalization, numerical construction, and edge cases. Each $96\times108\times96$ volume is divided into $12^3$-voxel patches, producing an $8\times9\times8$ grid with $N=576$. Average-pooling a FastSurfer region mask gives the fraction of each patch occupied by that region. Row normalization then changes the denominator from patch volume to region volume, so $C_{r,n}$ denotes the fraction of region $r$ contained in patch $n$. Consequently, $C\in\mathbb{R}^{R\times N}$ maps patch embeddings to $R$ region summaries as $(R\times N)(N\times D)\rightarrow(R\times D)$.

\paragraph{Worked Example.} Table~\ref{tab:coverage-example} illustrates a $4000$-voxel region spanning six patches. The normalized weights yield $z^{\mathrm{reg}}_{g,r}=0.225\,z_A+0.300\,z_B+0.200\,z_C+0.150\,z_D+0.100\,z_E+0.025\,z_F$; the other $570$ patches contribute zero. These values are illustrative rather than cohort measurements.

\begin{table}[h]
\floatconts
  {tab:coverage-example}
  {\caption{Illustrative construction of one row of $C$. The pooled values are fractions of each patch; the normalized weights are fractions of the region and sum to one.}}
  {\footnotesize\setlength{\tabcolsep}{5pt}
  \begin{tabular}{@{}lrrr@{}}
  \toprule
  Patch & Region voxels & Fraction of patch & $C_{r,n}$\\
  \midrule
  $A$ & $900$  & $0.52$ & $0.225$\\
  $B$ & $1200$ & $0.69$ & $0.300$\\
  $C$ & $800$  & $0.46$ & $0.200$\\
  $D$ & $600$  & $0.35$ & $0.150$\\
  $E$ & $400$  & $0.23$ & $0.100$\\
  $F$ & $100$  & $0.06$ & $0.025$\\
  \midrule
  Total & $4000$ & --- & $1.000$\\
  \bottomrule
  \end{tabular}}
\end{table}

\paragraph{Soft Regional Pooling.} Hard assignment to patch $B$ would discard the five patches containing $70\%$ of the example region, which is especially problematic for thin structures such as the entorhinal cortex and temporal horn. The weighted target preserves this distributed support, but remains a soft neighbourhood summary: a patch may receive high regional weight while containing substantial surrounding tissue. The attention losses therefore constrain where each token gathers information in addition to the regional content it predicts.

\paragraph{Implementation.} $C$ is recomputed for each scan from its segmentation and is sparse because each region intersects few patches. All-zero rows from absent labels are excluded from normalization and from region-averaged losses. Region targets are layer-normalized before loss computation.

Patch-level coverage averages boundary detail rather than requiring exact voxel boundaries, and masks are not needed for encoder inference. This motivates tolerance to small boundary errors but does not establish robustness to training-mask errors without a perturbation experiment.

\subsection{Components of \texttt{AG-JEPA}}
\label{app:components}

The Methods section defines the objectives; Table~\ref{tab:heads} instead summarizes the architectural role and failure mode addressed by each prediction component. Because anatomical and patch tokens share each encoder sequence, self-attention carries supervision across streams rather than making the objectives independently additive. The shared anatomy projector is introduced gradually and stabilized by the EMA target encoder and stop-gradient; attention alignment and overlap regularization additionally localize the tokens and discourage duplicate maps.

{Without $\mathcal{L}_{\mathrm{ap}}$, matching a pooled regional target need not retain masked-patch detail; without $\mathcal{L}_{\mathrm{pa}}$, cross-stream predictability is not enforced. The shared $\mathcal{L}_{\mathrm{ar}}$ projector anchors regional content, whereas $\mathcal{L}_{\mathrm{attn}}$ constrains its spatial source. These roles are complementary, not interchangeable.

\begin{table}[h] \floatconts {tab:heads} {\caption{Prediction components of \texttt{AG-JEPA} and their complementary architectural roles.}} {\scriptsize\setlength{\tabcolsep}{3pt}\renewcommand{\arraystretch}{0.95} \begin{tabular}{@{}llllp{0.47\textwidth}@{}} \toprule Component & Predicts & From & Enforces & Architectural role\\ \midrule $\mathcal{M}_{\mathrm{pp}}$ & $z^p_g$ & $z^p_c$ & image fidelity & Retains full-resolution patch detail and prevents collapse to regional summaries.\\ $\mathcal{M}_{\mathrm{ap}}$ & $z^p_g$ & $z^a_c$ & token content & Uses five tokens to predict $432$ masked patches, enforcing subject-specific content.\\ $\mathcal{M}_{\mathrm{pa}}$ & $z^a_g$ & $z^p_c$ & stream consistency & Predicts full-volume tokens from $144$ visible patches, closing the bidirectional stream loop.\\ $\mathcal{M}_{\mathrm{ar}}$ & $z^{\mathrm{reg}}_{g,r}$ & $z^a_{c,r}$ & region identity & Shares one projector across regions, placing identity in the tokens rather than per-region heads.\\ \bottomrule \end{tabular}} \end{table}

\newpage

\section{Dataset Details}
\label{app:data-details}

Scans are registered to the \href{https://www.mcgill.ca/bic/icbm152-152-nonlinear-atlases-version-2009}{MNI atlas} using \href{https://github.com/ANTsX/ANTs}{ANTs} and skull-stripped with \href{https://github.com/MIC-DKFZ/HD-BET}{HD-BET}. Diagnoses are assigned at each visit. Supervision includes age, MMSE, ADAS-13, CDR-SB, and MCI progression; converters are labelled pMCI, whereas subjects who do not convert within 36 months are labelled sMCI. Table~\ref{tab:data-composition} consolidates the cohort and longitudinal sample counts; endpoint totals vary with label availability. Sections~\ref{app:mrrate-labels} and~\ref{app:longitudinal-pairs} detail MR-RATE label extraction from radiology reports and longitudinal scan-pair construction, respectively.

\begin{table*}[htbp]
\floatconts
  {tab:data-composition}
  {\caption{Dataset composition. MRI scans are unique images across both visits; longitudinal tasks additionally report paired observations. Diagnosis-wise counts use visit-time labels and sum to the MRI-scan total. Merged-task rows report unique label-eligible samples across their listed endpoints.}}
  {\scriptsize
  \renewcommand{\arraystretch}{1.0}
  \begin{tabular*}{0.98\linewidth}{@{\extracolsep{\fill}}llrrrl@{}}
  \toprule
  \textbf{Use / endpoint} &
  \textbf{Cohort / split} &
  \textbf{MRI scans} &
  \textbf{Pairs} &
  \textbf{Subjects} &
  \shortstack{\textbf{MRI scan count}\\\textbf{diagnosis-wise}} \\
  \midrule
  \multicolumn{6}{@{}l}{\textit{\textbf{\textup{\texttt{AG-JEPA}} pretraining; CN/MCI--CN/AD classification; age/MMSE/ADAS-13/CDR-SB prediction}}} \\
  \addlinespace[1pt]
  \shortstack[l]{Pretraining + probe fitting\\Within-cohort evaluation} &
  \shortstack[l]{ADNI train\\ADNI Test} &
  \shortstack{2,523\\629} &
  \shortstack{--\\--} &
  \shortstack{540\\135} &
  \shortstack[l]{CN 737; MCI 1,027; AD 759\\CN 192; MCI 271; AD 166} \\
  Evaluation & MIRIAD & 708 & -- & 69 & CN 243; MCI 0; AD 465 \\
  Evaluation & MR-RATE & 220 & -- & 220 & CN 165; MCI 0; AD 55 \\
 Evaluation & AIBL & 1,023 & -- & 527 & CN 757; MCI 128; AD 138 \\
  \midrule
  \multicolumn{6}{@{}l}{\textit{\textbf{\textup{\texttt{AG-JEPA}}-based sMCI/pMCI classification and time-to-conversion prediction}}} \\
  \addlinespace[1pt]
  Probe fitting & ADNI train & 1,027 & -- & 269 & sMCI 692; pMCI 335 \\
  Held-out evaluation & ADNI Test & 271 & -- & 66 & sMCI 197; pMCI 74 \\
  \midrule
  \multicolumn{6}{@{}l}{\textit{\textbf{\textup{\texttt{Prog-AG-JEPA}}-based longitudinal progression (CN/AD and CN/MCI)}}} \\
  \addlinespace[1pt]
  Progression training + probe fitting & ADNI train & 2,288 & 4,373 & 477 & CN 676; MCI 942; AD 670 \\
  Future evaluation &
  ADNI Test &
  568 &
  1,070 &
  120 &
  CN 179; MCI 246; AD 143 \\
  \midrule
  \multicolumn{6}{@{}l}{\textit{\textbf{\textup{\texttt{Prog-AG-JEPA}}-based cognitive change (MMSE, ADAS-13, CDR-SB)}}} \\
  \addlinespace[1pt]
  Training + probe fitting & ADNI train & 2,283 & 4,358 & 477 & CN 675; MCI 940; AD 668 \\
  Evaluation & ADNI Test & 570 & 1,081 & 120 & CN 180; MCI 246; AD 144 \\
  \midrule
  \multicolumn{6}{@{}l}{\textit{\textbf{\textup{\texttt{Prog-AG-JEPA}}-based regional-volume change}}} \\
  \addlinespace[1pt]
  Training + probe fitting & ADNI train & 2,288 & 4,373 & 477 & CN 676; MCI 942; AD 670 \\
  Evaluation & ADNI Test & 572 & 1,087 & 120 & CN 180; MCI 248; AD 144 \\
  \bottomrule
  \end{tabular*}}
\end{table*}

\subsection{MR-RATE: Report-Derived Label Extraction}
\label{app:mrrate-labels}

We derive study-level labels for Alzheimer's disease (AD) from MR-RATE radiology reports using a deterministic rule-based classifier rather than a language model. Each label is traceable to the report spans that trigger the corresponding rules.

\paragraph{AD Mention Detection and Report Context.}
Reports are converted to lowercase and searched for the pattern \texttt{alzheimer('s)?}, the abbreviation \texttt{sdat}, and the phrase \texttt{senile dementia of alzheimer}. For each occurrence, we extract a context window extending 130 characters on either side and separately retain the 90 characters immediately preceding the mention. We also record whether the mention appears in the \emph{clinical indication}, which describes the referring physician's question or clinical history, or the \emph{report body}, comprising the findings and impression. This distinction separates a referral for suspected AD from a radiologist's assessment supporting AD.

\paragraph{Mention-Level Grading.}
Each mention is assigned one of four grades: \emph{exclusion}, \emph{assertion}, \emph{hedged assertion}, or \emph{hedge}, using the cues in Table~\ref{tab:mrrate-cues}. The assertion lexicon reflects the corpus's use of significance and compatibility language rather than exclusively explicit diagnostic statements.

\begin{table}[t]
\centering
\caption{Cue lexicons used to grade AD mentions in MR-RATE reports. These are mention-level grades; study-level labels are obtained by aggregation.}
\label{tab:mrrate-cues}
\small
\setlength{\tabcolsep}{5pt}
\renewcommand{\arraystretch}{1.15}
\begin{tabular}{@{}p{0.17\linewidth}p{0.78\linewidth}@{}}
\toprule
\textbf{Grade} & \textbf{Cues} \\
\midrule
Exclusion &
Clause-local negation: \emph{no}, \emph{not}, \emph{without},
\emph{negative for}, \emph{excludes}, \emph{ruled out},
\emph{unlikely}, and \emph{does not suggest/support/favor};
\emph{likelihood \ldots{} reduced/low}; \emph{largely excluded};
ERICA score 0--1. \\
\addlinespace
Assertion &
\emph{Significant for}, \emph{in terms of}, \emph{regarding},
\emph{consistent with}, \emph{compatible with}, \emph{supports},
\emph{favors}, \emph{suggests}, \emph{indicative of},
\emph{typical of}, \emph{diagnostic of}, \emph{primarily considered},
and \emph{likelihood \ldots{} high}; ERICA score 2--3. \\
\addlinespace
Hedged assertion &
A qualifier directly modifying an assertion cue, such as
\emph{may be significant}, \emph{potentially compatible}, or
\emph{possibly consistent}; alternatively, the diagnosis is
expressed as a question, such as \emph{Alzheimer's?}. \\
\addlinespace
Hedge &
\emph{Suspicious}, \emph{suspected}, \emph{possible},
\emph{probable}, \emph{may represent}, \emph{might},
\emph{could be}, \emph{cannot be excluded},
\emph{differential}, and \emph{versus}. \\
\bottomrule
\end{tabular}
\end{table}

\paragraph{Rule Precedence and Negation Scope.}
Cues are evaluated in the fixed order of exclusion, hedged assertion, assertion, and hedge. Testing hedged assertions before assertions prevents qualified statements, such as \emph{may be significant for Alzheimer's disease}, from being treated as unqualified assertions. Before exclusion testing, the phrase \emph{cannot be excluded} is masked because it expresses possibility rather than negation. Negation is applied only when the cue and disease mention occur within the same clause, without an intervening sentence or clause boundary. Thus, a statement such as \emph{there is no cortical infarct; the marked hippocampal atrophy supports Alzheimer's disease} is not classified as an AD exclusion. The construction \emph{not only \ldots{} but also} is likewise exempted from negation matching.

\paragraph{Selection of Positive and Negative Cases}
We considered Definitive studies as AD-positive and Excluded studies as AD-negative. Reports labelled Hedged or Indication Only represent uncertain or referral-only mentions of AD and were not treated as positive cases.

\paragraph{Study-Level Aggregation.}
For studies with AD mentions in the report body, the study is labelled \emph{definitive} if at least one mention is an assertion, \emph{hedged} if no mention is an assertion but at least one is a hedged assertion or hedge, and \emph{excluded} if all body mentions are exclusions. When AD appears only in the clinical indication, the study is labelled \emph{indication only}, unless the indication documents an established diagnosis rather than a diagnostic question. Established diagnoses are identified through patterns such as \emph{known}, \emph{history of}, \emph{previously diagnosed}, or \emph{patient with \ldots{} Alzheimer's}; these studies are assigned the \emph{definitive} label on clinical-history grounds. Here, \emph{definitive} denotes a report-derived category and does not imply independent clinical confirmation of AD.

\paragraph{Manual Review and Traceability.}
An author manually reviewed all 108 mention-containing reports alongside their assigned labels and extracted evidence spans. Of the 55 studies labelled \emph{definitive}, 49 were supported by a radiologist's assertion and six by a documented prior diagnosis. Per-mention evidence spans are provided alongside the labels to enable independent inspection of each decision.

\subsection{Longitudinal Scan-Pair Construction}
\label{app:longitudinal-pairs}

\paragraph{Unit of Analysis.}
The unit of analysis is an ordered within-subject scan pair $(s_a,s_b)$, comprising a current scan acquired at time $t_a$ and a future scan acquired at $t_b>t_a$, with follow-up interval $\Delta t=t_b-t_a$. All models in the longitudinal comparison, including \texttt{Prog-AG-JEPA} and the baselines, use identical training and test pair tables, ensuring that model comparisons are performed on matched samples.

\paragraph{Exhaustive Within-Subject Pairing.}
For a subject $i$ with $n_i$ eligible scans, we construct all forward-in-time scan combinations rather than restricting pairs to temporally adjacent visits. Before temporal filtering, the candidate pair counts are
\begin{linenomath}
\begin{equation}
\label{eq:longitudinal-pair-counts}
\begin{aligned}
P_i^{\mathrm{cand}}
    &= \binom{n_i}{2}
     = \frac{n_i(n_i-1)}{2}, \\
P_{\mathrm{total}}^{\mathrm{cand}}
    &= \sum_i P_i^{\mathrm{cand}}.
\end{aligned}
\end{equation}
\end{linenomath}
Each scan combination contributes one pair, with the earlier scan assigned as current; reversed pairs are not generated. This construction provides a broad range of observed follow-up intervals and supports evaluation at multiple horizons from a fixed baseline scan, such as $s_1\!\rightarrow\!s_2$, $s_1\!\rightarrow\!s_3$, and $s_1\!\rightarrow\!s_4$, using the same pair table.

\paragraph{Scan Eligibility.}
A scan is eligible only if it has a usable T1-weighted volume, a corresponding anatomical segmentation, and complete visit-date and age metadata. Both scans in a pair must satisfy these requirements. Subjects with fewer than two eligible scans contribute no longitudinal pairs.

\paragraph{Temporal Filtering.}
We retain only pairs with $\Delta t\geq180$ days (six months). This minimum interval is intended to reduce the influence of segmentation-related measurement variability on small longitudinal changes. No upper bound is imposed on $\Delta t$; the maximum retained interval is determined by the available observations.

\paragraph{Subject-Level Splitting.}
Training and test assignments are made at the subject level, and all pairs from a subject remain within the same split. Consequently, neither subjects nor scans are shared across the training and test sets. This prevents leakage that could arise from pair-level splitting, under which a scan appearing in multiple pairs could otherwise occur in both sets.

\paragraph{Statistical Dependence and Confidence Intervals.}
Each scan can participate in up to $n_i-1$ pairs, so pairs from the same subject are not treated as independent observations. Confidence intervals for longitudinal evaluations are therefore estimated using subject-level bootstrap resampling. Subjects are sampled with replacement, and all pairs belonging to each sampled subject are included together in the corresponding bootstrap replicate. This preserves within-subject dependence rather than resampling individual pairs independently.

\paragraph{Resulting Cohort (ADNI) and Interval Distribution.}
The longitudinal cohort comprises 597 subjects and 2,860 unique scans, yielding 6,296 candidate forward pairs. Applying the minimum-interval criterion retains 5,460 pairs: 4,373 training pairs from 477 subjects and 2,288 unique scans, and 1,087 test pairs from 120 subjects and 572 unique scans. There is no scan overlap between splits. Subjects contribute 2--13 usable scans and 1--64 retained pairs, with a median of eight pairs per subject. Follow-up intervals range from 0.50 to 4.38 years, with a median of 1.14 years. The interval groups of 0--1, 1--2, 2--3, and $>3$ years contain 1,764, 2,238, 1,010, and 448 pairs, respectively.

\section{Implementation Configuration}
\label{app:implementation}

Tables~\ref{tab:implementation} and~\ref{tab:prog-implementation} summarize the complete implementation sequence for \texttt{AG-JEPA} and \texttt{Prog-AG-JEPA}, respectively. Each table proceeds from input and masking specifications to model architecture, training configuration, and inference or downstream evaluation.
\paragraph{Probe selection.} A diagnosis-stratified, subject-level 25\% ADNI training hold-out selects feature pooling (anatomical tokens, brain-averaged patch tokens, or both) by validation AUROC. Test subjects are not used for selection, and AIBL receives the ADNI-fitted probe without adaptation.

\subsection{\texttt{AG-JEPA}}
\label{app:agjepa-implementation}

\begingroup
\scriptsize
\setlength{\tabcolsep}{2pt}
\renewcommand{\arraystretch}{1.08}
\begin{longtable}{@{}>{\raggedright\arraybackslash}p{0.165\textwidth}
                       >{\centering\arraybackslash}p{0.105\textwidth}
                       >{\raggedright\arraybackslash}p{0.175\textwidth}
                       @{\hspace{0.018\textwidth}}
                       >{\raggedright\arraybackslash}p{0.165\textwidth}
                       >{\centering\arraybackslash}p{0.105\textwidth}
                       >{\raggedright\arraybackslash}p{0.175\textwidth}@{}}
\caption{Key implementation configuration for \texttt{AG-JEPA}.}
\label{tab:implementation}\\
\toprule
\textbf{Parameter} & \textbf{Variable} & \textbf{Value} &
\textbf{Parameter} & \textbf{Variable} & \textbf{Value}\\
\midrule
\endfirsthead
\multicolumn{6}{c}{\tablename~\thetable\ continued}\\
\toprule
\textbf{Parameter} & \textbf{Variable} & \textbf{Value} &
\textbf{Parameter} & \textbf{Variable} & \textbf{Value}\\
\midrule
\endhead
\midrule
\multicolumn{6}{r}{Continued on next page}\\
\endfoot
\bottomrule
\endlastfoot

\multicolumn{6}{@{}l}{\textit{\textbf{Image and masking setup}}}\\
\addlinespace[2pt]
Input ROI & $\mathbf{x}$ & $96\times108\times96$ &
Patch size & $p$ & $12\times12\times12$\\
Patch tokens & $N$ & $576$ &
Context patch count & $N_c$ & $144$\\
Target patch count & $N_g$ & $432$ &
Mask blocks & $n_{\mathrm{blk}}$ & $16$\\
Target-mask ratio & $\rho_{\mathrm{mask}}$ & $0.75$ &
Mask aspect ratio & $a_{\mathrm{mask}}$ & $(0.75,1.5)$\\
Mask spatial scale & $s_{\mathrm{xy}}$ & $(0.2,0.5)$ &
Mask depth scale & $s_z$ & $(0.0,1.0)$\\
Foreground-aware masking & -- & true &
Region patch threshold & $\rho_{\mathrm{reg}}$ & $0.10$\\
Region context fraction & $\rho_{\mathrm{ctx}}$ & $0.25$ &
& & \\
\midrule
\multicolumn{6}{@{}l}{\textit{\textbf{Architecture}}}\\
\addlinespace[2pt]
Backbone & $f_{\theta}$ & 3D ViT-Base &
Embedding width & $D$ & $768$\\
Transformer blocks & $L$ & $12$ &
Position encoding & -- & prefix-aware RoPE\\
Attention kernel & -- & SDPA &
Regional prefix tokens & $R$ & $5$\\
Prefix initialization & $\sigma_{\mathrm{init}}$ & truncated normal, $0.02$ &
MoE blocks & $\mathcal{B}_{\mathrm{MoE}}$ & $\{1,3,5,7,9,11\}$\\
MoE routing & -- & bias-routed &
Routing score & -- & softmax\\
Shared experts & $n_{\mathrm{shared}}$ & $2$ &
Routed experts & $n_{\mathrm{routed}}$ & $16$\\
Active experts & $k$ & $6$ &
Expert hidden width & $D_{\mathrm{expert}}$ & $384$\\
Routing scale & $\alpha_{\mathrm{route}}$ & $4.0$ &
Predictor width / FFN & $(D_{\mathrm{pred}},D_{\mathrm{ff}})$ & $(384,1536)$\\
Predictor depth / heads & $(L_{\mathrm{pred}},H_{\mathrm{pred}})$ & $(4,12)$ &
Patch-to-patch head & $\mathcal{M}_{\mathrm{pp}}$ &
ViT-style Transformer; $144$ context patches and $432$ positioned patch queries with $L_{pred}$ and $H_{pred}$\\
Anatomy-to-patch head & $\mathcal{M}_{\mathrm{ap}}$ &
ViT-style Transformer; $5$ anatomy tokens and $432$ patch queries with $L_{pred}$ and $H_{pred}$&
Patch-to-anatomy head & $\mathcal{M}_{\mathrm{pa}}$ &
ViT-style Transformer; $144$ positioned context patches and $5$ anatomy queries with $L_{pred}$ and $H_{pred}$\\
Anatomy-to-region head & $\mathcal{M}_{\mathrm{ar}}$ &
Shared per-token MLP with GELU; $5$ region outputs &
& & \\
\midrule
\multicolumn{6}{@{}l}{\textit{\textbf{Training}}}\\
\addlinespace[2pt]
Optimizer & -- & AdamW &
Learning rate & $\eta$ & $3\times10^{-5}$, constant\\
Weight decay & $\lambda_{\mathrm{wd}}$ & $1\times10^{-5}$ &
Adam moments & $(\beta_1,\beta_2)$ & $(0.9,0.999)$\\
Gradient clipping & $g_{\max}$ & max-norm $1.0$ &
Training epochs & $E$ & $300$\\
Per-GPU batch size & $B_{\mathrm{gpu}}$ & $16$ &
Number of GPUs & $n_{\mathrm{gpu}}$ & $4$\\
Global batch size & $B$ & $64$ &
Training precision & -- & bfloat16 autocast\\
Random seed & $s$ & $42$, rank-offset &
Distributed training & -- & DDP\\
Target update & $\theta_{\mathrm{tgt}}$ & EMA each step &
EMA momentum & $m$ & $0.99925$\\
Target MoE biases & -- & copied from context &
Routing-bias rate & $\eta_{\mathrm{bias}}$ & $1\times10^{-4}$\\
Routing-bias clip & $b_{\max}$ & $0.3$ &
Predictive loss & $\ell_{\mathrm{pred}}$ & smooth L1\\
$\mathcal{M}_{\mathrm{pp}}/\mathcal{M}_{\mathrm{ap}}/\mathcal{M}_{\mathrm{pa}}$ weights & $\lambda_{\mathrm{pp,ap,pa}}$ & $1.0$ &
$\mathcal{M}_{\mathrm{ar}}$ weight & $\lambda_{\mathrm{ar}}$ & $1.0$, ramped\\
Attention localization & $\lambda_{\mathrm{attn}}$ & $0.5$, ramped &
Attention overlap & $\lambda_{\mathrm{ov}}$ & $0.01$\\
Ramp schedule & $w(t)$ & cosine &
Ramp minimum & $w_{\min}$ & $0.1$\\
Ramp maximum & $w_{\max}$ & $1.0$ &
Ramp start & $E_{\mathrm{start}}$ & epoch $20$\\
Ramp end & $E_{\mathrm{end}}$ & epoch $200$ &
Target normalization & -- & layer normalization\\
Region-loss validity & -- & present regions only &
& & \\
\midrule
\multicolumn{6}{@{}l}{\textit{\textbf{Inference and evaluation}}}\\
\addlinespace[2pt]
Feature poolings & $\mathcal{P}$ & anatomy, foreground, combined &
Probe estimator & -- & balanced logistic regression\\
Model-selection score & -- & AUROC &
Bootstrap resamples & $N_{\mathrm{boot}}$ & $1{,}000$\\
\end{longtable}
\endgroup

\subsection{\texttt{Prog-AG-JEPA}}
\begingroup
\scriptsize
\setlength{\tabcolsep}{2pt}
\renewcommand{\arraystretch}{1.08}
\begin{longtable}{@{}>{\raggedright\arraybackslash}p{0.165\textwidth}
                       >{\centering\arraybackslash}p{0.105\textwidth}
                       >{\raggedright\arraybackslash}p{0.175\textwidth}
                       @{\hspace{0.018\textwidth}}
                       >{\raggedright\arraybackslash}p{0.165\textwidth}
                       >{\centering\arraybackslash}p{0.105\textwidth}
                       >{\raggedright\arraybackslash}p{0.175\textwidth}@{}}
\caption{Key implementation configuration for \texttt{Prog-AG-JEPA}.}
\label{tab:prog-implementation}\\
\toprule
\textbf{Parameter} & \textbf{Variable} & \textbf{Value} &
\textbf{Parameter} & \textbf{Variable} & \textbf{Value}\\
\midrule
\endfirsthead

\multicolumn{6}{c}{\tablename~\thetable\ continued}\\
\toprule
\textbf{Parameter} & \textbf{Variable} & \textbf{Value} &
\textbf{Parameter} & \textbf{Variable} & \textbf{Value}\\
\midrule
\endhead

\midrule
\multicolumn{6}{r}{Continued on next page}\\
\endfoot

\bottomrule
\endlastfoot

\multicolumn{6}{@{}l}{\textit{\textbf{Input representation and conditioning}}}\\
\addlinespace[2pt]

Encoder & $f_{\theta}$ & \texttt{AG-JEPA} (\texttt{anatomy5}) &
Encoder status & -- & frozen\\

Patch tokens & $z_t$ & $576\times768$ &
Anatomy tokens & $a_t$ & $5\times768$\\

Conditioning variables & $c_t$ &
$\Delta t$, age, regional volumes &
Conditioning dimension & $d_c$ & $7$\\

Conditioning injection & -- & learned prefix &
Prefix tokens & $N_{\mathrm{cond}}$ & $2$\\

\midrule
\multicolumn{6}{@{}l}{\textit{\textbf{Architecture}}}\\
\addlinespace[2pt]

Predictor & $\mathcal{P}_t$ & transformer &
Predictor depth & $L_{\mathrm{pred}}$ & $4$\\

Predictor width & $D_{\mathrm{pred}}$ & $384$ &
Attention heads & $H_{\mathrm{pred}}$ & $6$\\

Predictor input & -- &
patch $+$ anatomy tokens &
Conditioning input & -- & prefix from $c_t$\\

Predicted patch drift & $\Delta\hat{z}_t$ & $576\times768$ &
Predicted anatomy drift & $\Delta\hat{a}_t$ & $5\times768$\\

Future patch tokens & $\hat{z}_{t+\Delta t}$ &
$z_t+\Delta\hat{z}_t$ &
Future anatomy tokens & $\hat{a}_{t+\Delta t}$ &
$a_t+\Delta\hat{a}_t$\\

\midrule
\multicolumn{6}{@{}l}{\textit{\textbf{Training}}}\\
\addlinespace[2pt]

Optimizer & -- & AdamW &
Learning rate & $\eta$ & $3\times10^{-4}$\\

Weight decay & $\lambda_{\mathrm{wd}}$ & $0.05$ &
Gradient clipping & $g_{\max}$ & max-norm $1.0$\\

Training epochs & $E$ & $30$ &
Batch size & $B$ & $32$\\

Cross-validation folds & -- & $3$ &
Seed ensemble & -- & $6$ seeds, averaged\\

Patch-drift target & $\Delta z_t$ &
$z_{t+\Delta t}-z_t$ &
Anatomy-drift target & $\Delta a_t$ &
$a_{t+\Delta t}-a_t$\\

Patch loss weight & $\lambda_z$ & $1.0$ &
Anatomy loss weight & $\lambda_a$ & $0.3$\\

Loss & $\mathcal{L}_{pt}$ &
weighted smooth L1 &
Smooth-L1 parameter & $\beta$ & $1.0$\\

Loss reduction & -- &
mean over features and tokens &
Target normalization & -- &
per dimension, training data only\\

\midrule
\multicolumn{6}{@{}l}{\textit{\textbf{Inference and downstream evaluation}}}\\
\addlinespace[2pt]

Cognitive/volume regression & -- & ridge regression &
Regression input & -- & drift tokens\\

Drift readout & -- & anatomy tokens $\|$ foreground pool &
Classification input & -- & predicted future latents\\

Regression standardization & -- & training statistics &
Bootstrap resamples & $N_{\mathrm{boot}}$ & $2{,}000$, subject-level\\






\end{longtable}
\endgroup

\section{Cross-Cohort Representation Analysis}
\label{app:representation-generalization}

\begin{figure}[ht]
\centering
\includegraphics[width=\textwidth]{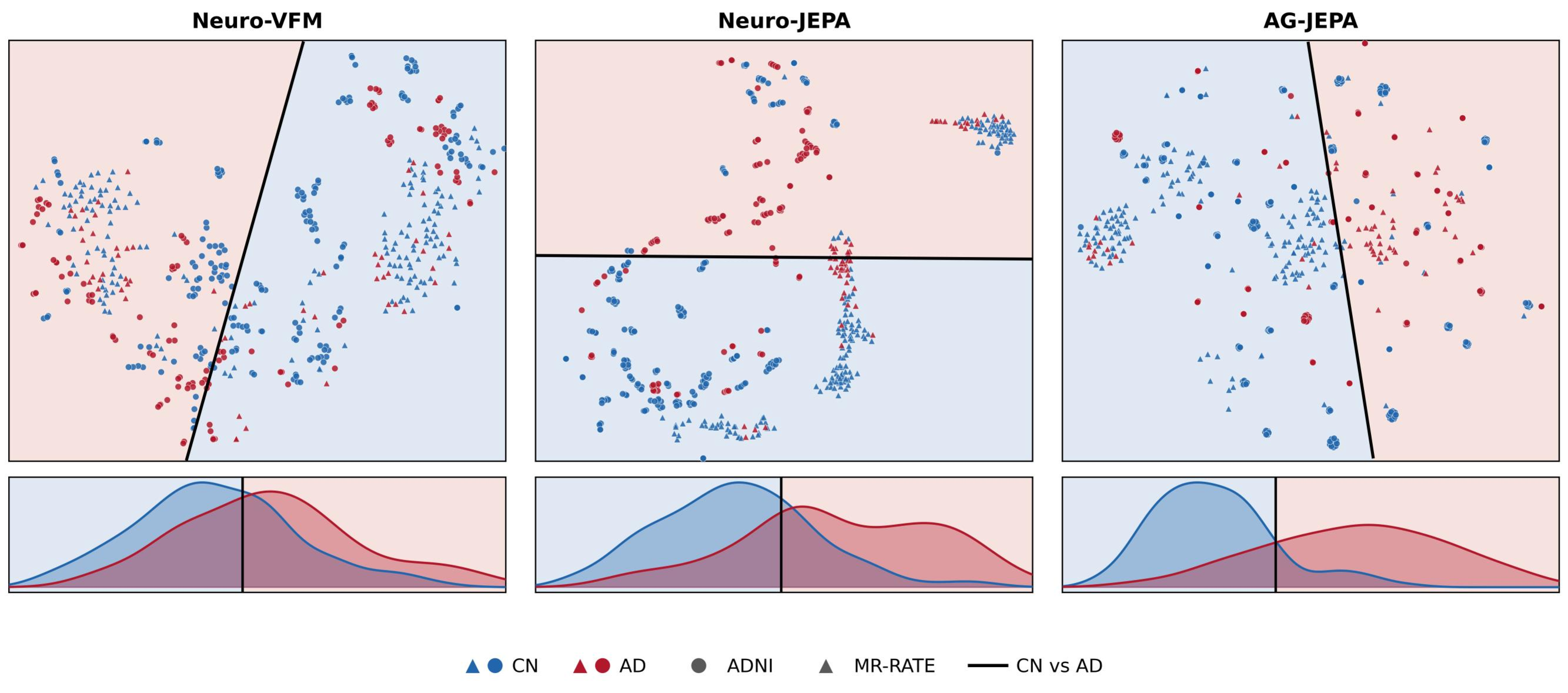}
\caption{Cross-cohort CN-versus-AD latent organization for ADNI Test (circles) and MR-RATE (triangles), with the ADNI train-fitted linear decision boundary (black) and discriminant-score distributions (bottom). \texttt{AG-JEPA} retains clearer diagnostic separation across cohorts, while \texttt{NeuroVFM} and \texttt{Neuro-JEPA} show greater cohort-dependent clustering and class overlap.}
\label{fig:qual-diag}
\end{figure}

\begin{figure}[ht]
\centering
\includegraphics[width=\textwidth]{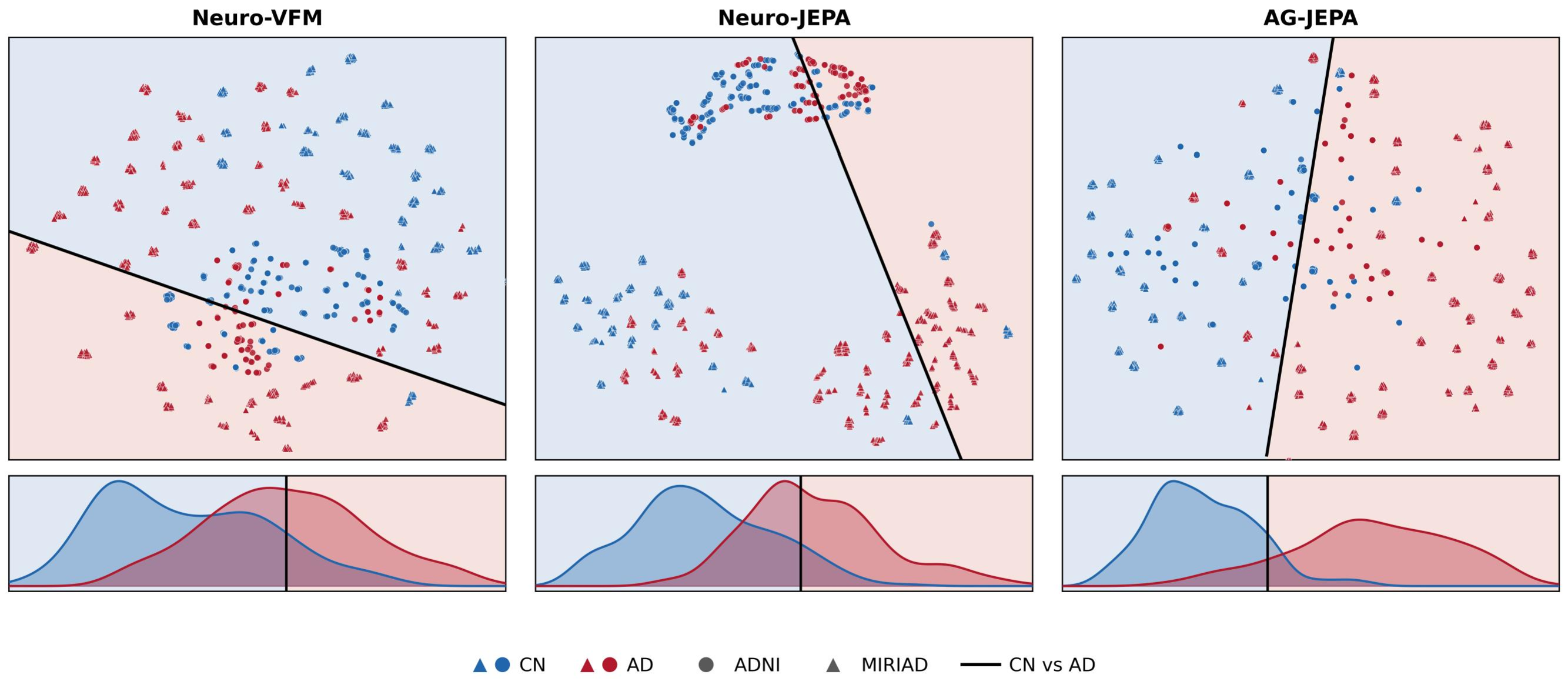}
\caption{Cross-cohort CN-versus-AD latent organization for ADNI Test (circles) and MIRIAD (triangles), with the ADNI train-fitted linear decision boundary (black) and discriminant-score distributions (bottom).}
\label{fig:qual-diag-miriad}
\end{figure}

Figures~\ref{fig:qual-diag} and~\ref{fig:qual-diag-miriad} show that the ADNI train-fitted boundary transfers more consistently to MR-RATE and MIRIAD for \texttt{AG-JEPA}: CN and AD remain organized primarily by diagnosis rather than cohort, with less overlap in their discriminant distributions. The baselines instead form stronger cohort-specific clusters, indicating relatively poorer preservation of disease-related structure under cohort shift.

\begingroup

Complementing MR-RATE and MIRIAD, Figure~\ref{fig:revision-aibl-diagnosis} extends this analysis to unseen AIBL for CN/MCI and CN/AD. It provides qualitative evidence of relatively better diagnostic separation for \texttt{AG-JEPA}, not a statistical test of cohort invariance.

\begin{figure}[htbp]

\centering
\includegraphics[width=0.75\textwidth]{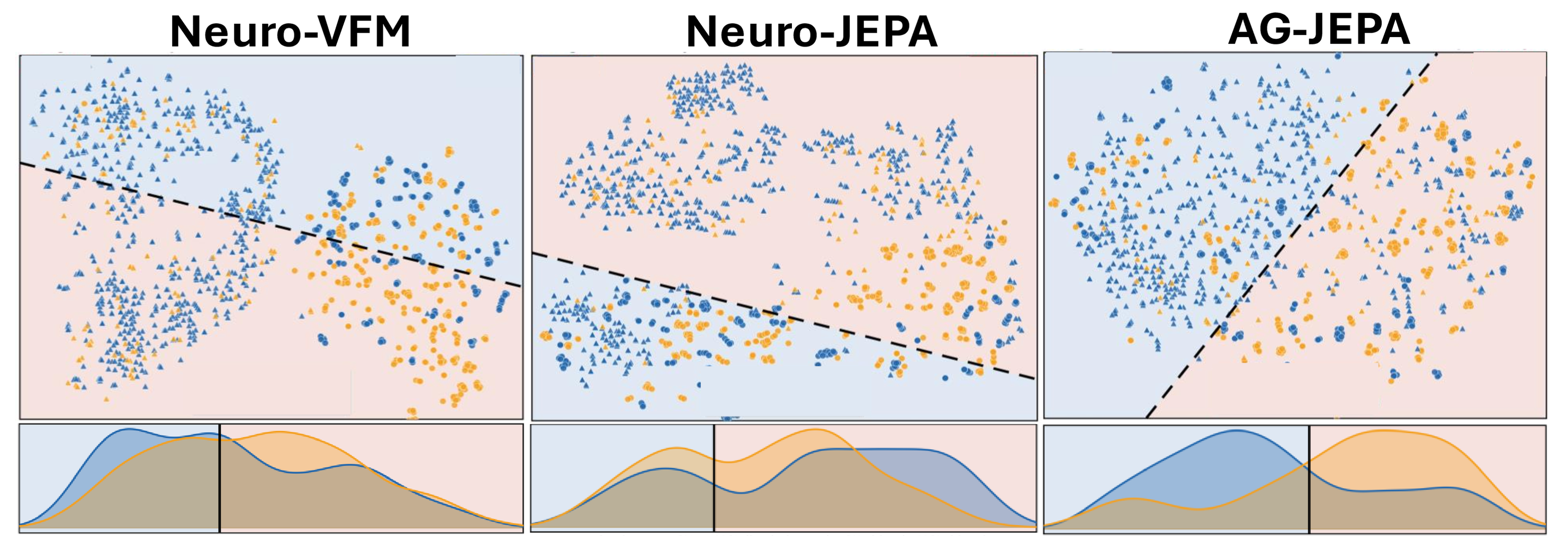}\\[4pt]
\includegraphics[width=0.75\textwidth]{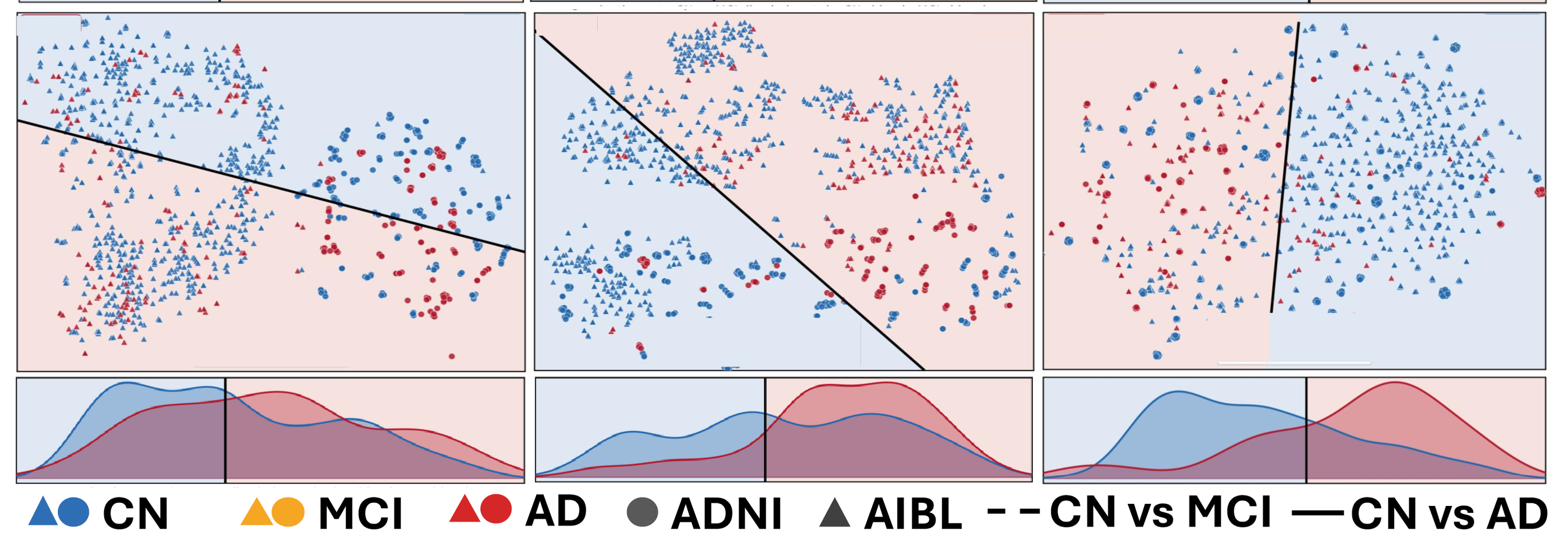}
\caption{ADNI Test and unseen AIBL diagnostic organization with ADNI Train-fitted discriminants: CN/MCI (top) and CN/AD (bottom).}
\label{fig:revision-aibl-diagnosis}
\end{figure}
\endgroup

\clearpage
\section{Regional Volume Evaluation}
\label{app:regional-volume-eval}
    For \texttt{Prog-AG-JEPA}, ridge regression estimates regional volume changes directly from predicted patch and anatomical latent drifts, without synthesizing follow-up images. For \texttt{BrLP}, \texttt{Latent Drift}, and \texttt{AG-LDM}, we use their native SynthSeg outputs to obtain regional volumes from generated follow-up scans and compute changes relative to current-scan volumes. We exclude the entorhinal cortex from these comparisons because the native SynthSeg outputs used for the baselines do not provide separate measurements for this region.

For region $r$, the ground-truth relative volume change between the current and follow-up scans is
\begin{equation}
\Delta V_{t,r}
=
\frac{V_{t+\Delta t,r}}{V_{t,r}}-1,
\label{eq:regional-relative-change}
\end{equation}
and $\widehat{\Delta V}_{t,r}$ denotes the corresponding method-specific estimate of relative volume change. All methods are evaluated using
\begin{equation}
\mathrm{MAE}_{r}^{\%\mathrm{TBV}}
=
\frac{100}{N}\sum_{t=1}^{N}
\left|\widehat{\Delta V}_{t,r}-\Delta V_{t,r}\right|
\frac{V_{t,r}}{\mathrm{TBV}_t}.
\label{eq:regional-volume-mae}
\end{equation}
Here, $N$ is the number of evaluated scan pairs and $\mathrm{TBV}_t$ is the total brain volume at the current scan.

\begin{figure}[t]
  \centering
  \includegraphics[width=\linewidth]{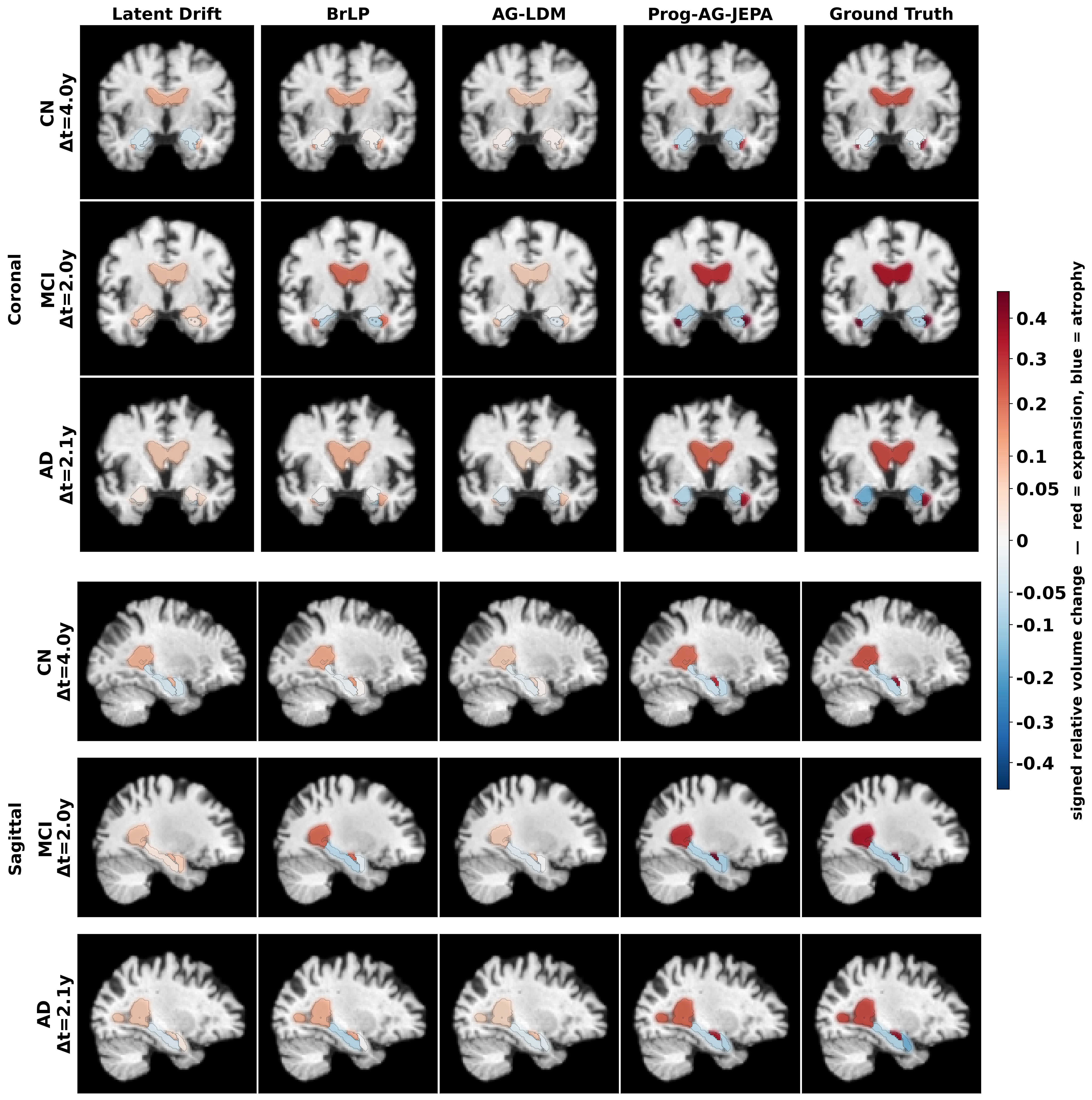}
  \caption{Signed relative regional volume change $\Delta V_{t,r}$, shown in coronal
  (top) and sagittal (bottom) views. Rows are illustrative scan pairs whose class is the
  diagnosis at the current (baseline) scan; $\Delta t$ is the interval to the follow-up
  scan. All panels share a single colour bar (red, expansion; blue, atrophy), clipped at
  $\pm0.47$.}
  \label{fig:volchange}
\end{figure}

Eq.~\eqref{eq:regional-volume-mae} weights each region's relative error by its share of total brain volume, $V_{t,r}/\mathrm{TBV}_t$. This places errors from regions of different sizes on a common scale and makes them comparable across regions of very different sizes and cohorts with differing head sizes. Hence, a structure occupying a small fraction of
the brain therefore contributes proportionally to the reported error, and the metric
cannot be inflated by large relative fluctuations in small structures.

Figure~\ref{fig:volchange} visualises the signed relative change $\Delta V_{t,r}$ of Eq.~\eqref{eq:regional-relative-change} for the ground truth and model predictions before TBV weighting. Positive values (red) indicate expansion and negative values (blue) indicate atrophy, with all columns shown on the same scale. For example, $\Delta V_{t,r}=-0.2$ indicates a 20\% decrease in the region's volume relative to the current scan.

Across both views, \texttt{Prog-AG-JEPA} more closely reproduces the reference pattern, including ventricular and temporal-horn expansion alongside medial-temporal atrophy. In contrast, the comparison methods often underestimate the magnitude of regional change, producing predictions closer to zero, and occasionally predict the incorrect direction of change.

\begingroup
\subsection{Matched Longitudinal Readouts}
\label{app:longitudinal-controls}

To assess the native readout choice described above, we fit the same ridge readout to the generative models' latent changes on identical training pairs. On pooled ADNI+AIBL test pairs, generated-image segmentation is relatively better for BrLP and Latent Drift and numerically comparable for AG-LDM (Table~\ref{tab:revision-matched-readout}). This supports the native readout used for these baselines; our model has the lowest MAE under either comparison. Matching the regression readout does not make the models' representations or training objectives identical.

\begin{table}[htbp]
\floatconts
{tab:revision-matched-readout}
{\caption{Volume-change MAE (\%TBV) on pooled ADNI+AIBL pairs, averaged over hippocampus, amygdala, temporal horn and lateral ventricle. Native: generated-image segmentation with SynthSeg; ridge: latent regression.}}
{\small
\begin{tabular}{@{}lcc@{}}
\toprule
Model & Native MAE & Ridge MAE \\
\midrule
\texttt{Prog-AG-JEPA} & N/A & 0.06503 \\
BrLP & 0.07184 & 0.07606 \\
Latent Drift & 0.08539 & 0.09297 \\
AG-LDM & 0.08182 & 0.08162 \\
\bottomrule
\end{tabular}}
\end{table}

Current-state and covariate-only ablations are reported in Appendix~\ref{app:current-state-ablation}.

\endgroup
\clearpage

\section{Additional Ablations}
\label{app:prog-conditioning}

\begingroup
\subsection{Anatomical Guidance versus Spatial Supervision}
\label{app:anatomical-controls}

The spatial-control JEPA adds a VoCo-inspired cosine loss~\citep{wu2026voco} that matches sub-volume embedding similarity to spatial overlap, without FastSurfer masks or anatomical tokens. It uses the same backbone, data and training schedule. \texttt{AG-JEPA} is relatively better across the evaluated tasks (Table~\ref{tab:revision-spatial-control}), supporting the full anatomy-guided design beyond spatial awareness alone; the comparison does not isolate tokens from their accompanying objectives.

\begin{table}[htbp]
\floatconts
{tab:revision-spatial-control}
{\caption{Spatial-control JEPA versus \texttt{AG-JEPA} (AUROC). ADNI-fitted probes are evaluated without AIBL adaptation.}}
{\small
\begin{tabular}{@{}llcc@{}}
\toprule
Dataset & Task & Spatial-control JEPA & \texttt{AG-JEPA} \\
\midrule
ADNI & CN/AD & 0.875 & 0.887 \\
ADNI & CN/MCI & 0.742 & 0.804 \\
AIBL & CN/AD & 0.856 & 0.905 \\
AIBL & CN/MCI & 0.631 & 0.705 \\
\bottomrule
\end{tabular}}
\end{table}
\endgroup

\subsection{Conditioning Variables and Anatomical Tokens in \texttt{Prog-AG-JEPA}}
\label{app:conditioning-ablation}
Table~\ref{tab:prog-cond} evaluates predictor conditioning and the contribution of anatomical tokens in \texttt{Prog-AG-JEPA}. Conditioning on time, age, and regional volumes consistently improves the diagnostic, cognitive, and volume-change estimates over using time alone or time with age, supporting their joint use in the progression head. With all three conditioning variables retained, restricting the predictor to patch-token inputs and outputs yields poorer performance overall.
\begin{table*}[!h]
\floatconts
  {tab:prog-cond}
  {\caption{\texttt{Prog-AG-JEPA} ablation on anatomical tokens and predictor conditioning using ADNI test pairs.
  Classification is AUROC ($\uparrow$); cognitive-score change and regional-volume change (\%TBV) are MAE ($\downarrow$).}}
  {\small
  \setlength{\tabcolsep}{4pt}
  \resizebox{0.99\textwidth}{!}{%
  \begin{tabular}{clcccccc}
  \toprule
  \multirow{2}{*}{\shortstack{Anatomical\\tokens}}
& \multirow{2}{*}{\shortstack{Predictor\\conditioning}}
& \multicolumn{2}{c}{AUROC}
& \multicolumn{3}{c}{Cognitive Score Change (MAE)}
& \multicolumn{1}{c}{Volume (MAE)} \\
\cmidrule(lr){3-4}
\cmidrule(lr){5-7}
\cmidrule(lr){8-8}
&
& CN/AD$\uparrow$
& CN/MCI$\uparrow$
& MMSE$\downarrow$
& ADAS-13$\downarrow$
& CDR-SB$\downarrow$
& \%TBV$\downarrow$ \\
  \midrule

  \multirow{3}{*}{Yes}
  & $\Delta t$
  & $0.836{\pm}0.054$
  & $0.744{\pm}0.065$
  & $1.939{\pm}0.159$
  & $4.299{\pm}0.231$
  & $1.018{\pm}0.064$
  & $0.064{\pm}0.005$ \\

  & $\Delta t$ + age
  & $0.837{\pm}0.054$
  & $0.746{\pm}0.065$
  & $1.921{\pm}0.160$
  & $4.275{\pm}0.231$
  & $1.000{\pm}0.063$
  & $0.062{\pm}0.004$ \\

  \rowcolor{agrow}
  & $\Delta t$ + age + vol.
  & $\mathbf{0.844{\pm}0.052}$
  & $\mathbf{0.765{\pm}0.060}$
  & $\mathbf{1.893{\pm}0.159}$
  & $4.246{\pm}0.230$
  & $\mathbf{0.911{\pm}0.064}$
  & $\mathbf{0.059{\pm}0.004}$ \\

  \midrule

  No
  & $\Delta t$ + age + vol.
  & $0.822{\pm}0.059$
  & $0.740{\pm}0.065$
  & $1.897{\pm}0.154$
  & $\mathbf{4.200{\pm}0.223}$
  & $0.976{\pm}0.058$
  & $0.064{\pm}0.005$ \\

  \bottomrule
  \end{tabular}}}
\end{table*}

\begingroup

\subsection{Single-Drift-Anatomical Token Analysis in \texttt{Prog-AG-JEPA}}
\label{app:token-validation}


Table~\ref{tab:revision-drift-probes} applies the same ridge readout to one predicted anatomical drift token at a time on ADNI Test. Individual tokens retain progression information, but the displayed errors do not establish that each token preferentially predicts its own region. The entorhinal endpoint is included only for our model because the baselines' SynthSeg outputs do not provide it.

\begin{table}[htbp]
\floatconts
{tab:revision-drift-probes}
{\caption{Regional volume-change MAE (\%TBV) from single drift tokens on ADNI Test. Full readout: all anatomical drift tokens and brain-averaged patch drift.}}
{\small
\begin{tabular}{@{}lcccccc@{}}
\toprule
Readout & Hipp. & Amyg. & Ent. & Temp.\ horn & Lat.\ vent. & Mean, five regions \\
\midrule
Full & 0.018 & 0.010 & 0.015 & 0.024 & 0.185 & 0.0505 \\
Hippocampus & 0.018 & 0.010 & 0.015 & 0.025 & 0.194 & 0.0524 \\
Amygdala & 0.017 & 0.010 & 0.015 & 0.025 & 0.200 & 0.0535 \\
Entorhinal cortex & 0.018 & 0.010 & 0.015 & 0.025 & 0.192 & 0.0520 \\
Temporal horn & 0.018 & 0.010 & 0.015 & 0.025 & 0.200 & 0.0536 \\
Lateral ventricle & 0.018 & 0.010 & 0.014 & 0.025 & 0.200 & 0.0534 \\
\bottomrule
\end{tabular}}
\end{table}

\subsection{Current-State and Covariate-Only Ablation for \texttt{Prog-AG-JEPA}}
\label{app:current-state-ablation}

Table~\ref{tab:revision-current-covariates} separately evaluates the ADNI Test ablations. The covariate-only baseline uses $c_t=(\Delta t,\mathrm{Age}_t,\mathrm{Vol}_t)$; the current-state baseline adds current imaging features but no predicted drift; the full model additionally predicts latent progression. Imaging features primarily improve diagnosis, while latent progression further improves all displayed endpoints over the full current state.

\begin{table}[htbp]
\floatconts
{tab:revision-current-covariates}
{\caption{Current-state and covariate-only controls on ADNI Test. Cognitive MAE averages MMSE, ADAS-13 and CDR-SB on their original scales; volume-change MAE (\%TBV) averages the four shared regions.}}
{\small
\begin{tabular}{@{}lcccc@{}}
\toprule
& \multicolumn{2}{c}{Diagnosis (AUROC)} & Cognitive change (MAE) & Volume change (MAE) \\
\cmidrule(lr){2-3}
Model & CN/AD & CN/MCI & Mean & Mean \\
\midrule
Covariate-only & 0.5864 & 0.5952 & 2.4982 & 0.07255 \\
Current-state & 0.8291 & 0.7521 & 2.4160 & 0.07181 \\
\texttt{Prog-AG-JEPA} & 0.8435 & 0.7648 & 2.3500 & 0.05939 \\
\bottomrule
\end{tabular}}
\end{table}

\section{Statistical Significance}
\label{app:paired-validation}

We compare diagnostic AUROCs with two-sided paired DeLong tests on the same test subjects (Table~\ref{tab:revision-diagnosis}). All seven comparisons have $p<0.05$. These tests concern the listed endpoints, not all downstream tasks or ablation differences. The paired results retain their original evaluation configurations.


\begin{table}[htbp]
\floatconts
{tab:revision-aibl-summary}
{\caption{AIBL CN/AD AUROC for the cohort-summary configurations shown in Figure~\ref{fig:mrrate-cnad}. Probes are fitted on ADNI Train and evaluated on 449 AIBL subjects.}}
{\small
\begin{tabular}{@{}lcc@{}}
\toprule
Method & AUROC & 95\% bootstrap CI \\
\midrule
\texttt{NeuroVFM} & $0.790\pm0.028$ & [0.734, 0.839] \\
\texttt{Neuro-JEPA} & $0.892\pm0.019$ & [0.851, 0.927] \\
\texttt{AG-JEPA} & $0.914\pm0.018$ & [0.876, 0.947] \\
\bottomrule
\end{tabular}}
\end{table}

\begin{table}[htbp]
\floatconts
{tab:revision-diagnosis}
{\caption{Paired diagnostic comparison. $\Delta$: \texttt{AG-JEPA} minus \texttt{Neuro-JEPA}; CI: 95\% DeLong confidence interval.}}
{\small
\begin{tabular}{@{}llcccc@{}}
\toprule
Dataset & Task & \texttt{Neuro-JEPA} & \texttt{AG-JEPA} & $\Delta$ [95\% CI] & $p$ \\
\midrule
ADNI & CN/MCI & 0.720 & 0.804 & $+0.084$ [$+0.040$, $+0.128$] & 0.0002 \\
ADNI & CN/AD & 0.876 & 0.887 & $+0.011$ [$+0.005$, $+0.017$] & 0.0001 \\
ADNI & sMCI/pMCI & 0.628 & 0.689 & $+0.061$ [$+0.007$, $+0.115$] & 0.0257 \\
MR-RATE & CN/AD & 0.875 & 0.946 & $+0.071$ [$+0.040$, $+0.102$] & $6.9\times10^{-6}$ \\
MIRIAD & CN/AD & 0.966 & 0.981 & $+0.015$ [$+0.008$, $+0.022$] & $1.2\times10^{-5}$ \\
AIBL & CN/MCI & 0.661 & 0.705 & $+0.044$ [$+0.019$, $+0.069$] & 0.0006 \\
AIBL & CN/AD & 0.892 & 0.914 & $+0.034$ [$+0.015$, $+0.053$] & 0.0005 \\
\bottomrule
\end{tabular}}
\end{table}

Table~\ref{tab:revision-progression-tests} separately reports pooled ADNI+AIBL longitudinal comparisons, not the ADNI-only results of Table~\ref{tab:prog-cog}. Diagnosis uses paired DeLong tests; volume-change MAE uses a paired subject-level bootstrap with 2,000 resamples. The lateral-ventricle difference is not significant at $\alpha=0.05$.

\begin{table}[htbp]
\floatconts
{tab:revision-progression-tests}
{\caption{{Paired longitudinal comparison on ADNI+AIBL. $\Delta$: \texttt{Prog-AG-JEPA} minus BrLP, with 95\% CI. Volume-change MAE is in \%TBV.}}}
{\small
\begin{tabular}{@{}lcccc@{}}
\toprule
Task & BrLP & \texttt{Prog-AG-JEPA} & $\Delta$ [95\% CI] & $p$ \\
\midrule
CN/AD (AUROC) & 0.870 & 0.886 & $+0.016$ [$+0.001$, $+0.031$] & 0.036 \\
CN/MCI (AUROC) & 0.745 & 0.763 & $+0.018$ [$+0.000$, $+0.036$] & 0.047 \\
Hippocampus (MAE) & 0.023 & 0.019 & $-0.004$ [$-0.00825$, $-0.00174$] & 0.0005 \\
Amygdala (MAE) & 0.015 & 0.012 & $-0.003$ [$-0.00903$, $-0.00016$] & 0.026 \\
Temporal horn (MAE) & 0.029 & 0.023 & $-0.006$ [$-0.00877$, $-0.00431$] & 0.0005 \\
Lateral ventricle (MAE) & 0.220 & 0.205 & $-0.015$ [$-0.03487$, $+0.00627$] & 0.178 \\
Mean, four regions (MAE) & 0.072 & 0.065 & $-0.007$ [$-0.01305$, $-0.00117$] & 0.015 \\
\bottomrule
\end{tabular}}
\end{table}

\endgroup

\end{document}